\documentclass{article}
\usepackage{arxiv_public,times}

\usepackage{amsmath,amsfonts,bm}

\def\eqref#1{equation~\ref{#1}}

\def\1{\bm{1}}

\def\vzero{{\bm{0}}}

\def\va{{\bm{a}}}

\def\vs{{\bm{s}}}

\def\vw{{\bm{w}}}
\def\vx{{\bm{x}}}

\def\vz{{\bm{z}}}

\def\mI{{\bm{I}}}

\def\mM{{\bm{M}}}

\def\mP{{\bm{P}}}

\def\mR{{\bm{R}}}
\def\mS{{\bm{S}}}

\def\mU{{\bm{U}}}
\def\mV{{\bm{V}}}
\def\mW{{\bm{W}}}

\def\mZ{{\bm{Z}}}

\DeclareMathAlphabet{\mathsfit}{\encodingdefault}{\sfdefault}{m}{sl}
\SetMathAlphabet{\mathsfit}{bold}{\encodingdefault}{\sfdefault}{bx}{n}

\newcommand{\Var}{\mathrm{Var}}

\definecolor{citecolor}{HTML}{0071bc}
\usepackage[pagebackref=false,breaklinks=true,letterpaper=true,colorlinks,citecolor=citecolor,bookmarks=false]{hyperref}
\usepackage{url}
\usepackage{graphicx}
\usepackage[utf8]{inputenc}
\usepackage[T1]{fontenc}
\usepackage{booktabs}
\usepackage{amsfonts}
\usepackage{nicefrac}
\usepackage{microtype}
\usepackage{xcolor}
\usepackage{amsmath}
\usepackage{amsthm}
\usepackage{subfigure}
\usepackage{multirow}
\usepackage{makecell}
\usepackage{pifont}
\usepackage{wrapfig}
\usepackage{threeparttable}
\usepackage{bm}
\usepackage{algorithm}
\usepackage[noend]{algorithmic}

\usepackage{thm-restate}

\declaretheorem[name=Theorem]{theorem}

\usepackage{enumitem}
\setlist[enumerate]{leftmargin=*}
\setlist[enumerate,1]{itemsep=0pt,topsep=0pt,parsep=0pt}

\allowdisplaybreaks[1]

\title{SubZero+: Memory-Efficient Adaptive Zeroth-Order LLM Fine-Tuning in Random Subspaces}

\author{
	Ziming Yu\textsuperscript{1},
	Shuyao Xiao\textsuperscript{1},
	Xingyu Zhao\textsuperscript{1},
	Sike Wang\textsuperscript{1},
	Pan Zhou\textsuperscript{2},
	Peiyu Zang\textsuperscript{1},	\\
	~\textbf{Xiangda Yan}\textsuperscript{\textbf{3}},
	\textbf{Yongjie Yang}\textsuperscript{\textbf{3}},
	\textbf{Jia Li}\textsuperscript{\textbf{1,4,5}}\thanks{Corresponding Author}~\\
	\textsuperscript{1~} Beijing Normal University \quad
	\textsuperscript{2~} Singapore Management University \quad
	\textsuperscript{3~} Xiaomi Inc. \quad \\
	\textsuperscript{4~} Beijing Key Laboratory of Artificial Intelligence for Education \quad \\
	\textsuperscript{5~} Engineering Research Center of Intelligent Technology and Educational Application (MOE) \\
	\texttt{\{zimingyu, shuyaoxiao, xingyuzhao, sikewang\}@mail.bnu.edu.cn} \\
	\texttt{panzhou@smu.edu.sg, jiali@bnu.edu.cn}
}

\begin{document}

\maketitle

\begin{abstract}
Zeroth-order (ZO) optimization with SGD in random subspaces enables memory-efficient fine-tuning of large language models without backpropagation. However, high gradient estimation noise fundamentally undermines adaptive optimizers like Adam. We propose SubZero+, which achieves practical adaptive ZO optimization through a carefully designed dual low-dimensionality strategy: (i) multi-query forward-difference gradient estimation in periodically refreshed random subspaces to mitigate noise amplification in moment buffers, and (ii) Adam updates with periodic restarts performed directly in low-dimensional space rather than full-parameter space. In experiments, this dual design retains memory overhead comparable to momentum-free ZO methods while achieving stronger optimization performance than the evaluated ZO baselines. Theoretically, in the exact-directional limit, $K$-query averaging preserves conditional unbiasedness, while the coefficient estimator's covariance and mean-squared error, as well as query-induced second-moment inflation, scale exactly as $1/K$. Extensive experiments across SuperGLUE with models from 1.3B to
32B parameters under both full fine-tuning and LoRA schemes demonstrate consistent improvements over competing ZO methods. SubZero+ significantly narrows the performance gap with first-order optimization while preserving ZO's inference-time memory efficiency.
\end{abstract}

\section{Introduction}
\label{sec: intro}
\vspace*{-0.5em}

Fine-tuning large language models (LLMs) is essential  for adapting pretrained models to downstream tasks. Most existing methods rely on backpropagation (BP) to compute first-order (FO) gradients for optimization~\citep{amari1993backpropagation,kingma2014adam}. As a principled alternative, zeroth-order (ZO) optimization enables fine-tuning without backward passes~\citep{malladi2023fine,zhang2024revisiting,liu2024sparse,yu2025zeroth}. By removing BP, ZO methods can significantly reduce activation-memory overhead, simplify the training pipeline, and make fine-tuning practical in scenarios where BP is infeasible or inefficient~\citep{liu2026position}. However, in practice it remains challenging to obtain stable and competitive performance with ZO optimization.

A promising direction combines ZO estimation with subspace restriction. In particular, SubZero~\citep{yu2025zeroth} projects ZO gradient estimates onto layer-wise low-dimensional subspaces that are periodically refreshed for both efficiency and effectiveness. This design is motivated by the observation that fine-tuning gradients are often approximately low-rank~\citep{zhang2023fine,malladi2023fine,zhao2024galore,hao2024flora}. By reducing the effective dimension, subspace methods can improve estimation quality while preserving the inference-level memory benefits of ZO training. Nonetheless, the residual estimation noise poses a particular challenge for adaptive optimizers such as Adam~\citep{kingma2014adam}, since the same noisy estimate is used in both the numerator and the denominator of the parameter update.
 
A critical observation reveals that despite gradient accumulation across multiple queries, existing methods achieve only marginal improvements in ZO LLM fine-tuning~\citep{peng2025muzo}. We pinpoint a fundamental bottleneck: by performing both gradient estimation and parameter updates in the full-parameter space, existing methods suffer from dimension-dependent variance that dominates even after averaging multiple queries~\citep{duchi2015optimal,nesterov2017random}. Consequently, variance reduction scales only as $O(1/K)$ for $K$ queries, while the dimension-dependent noise dominates. Furthermore, adaptive optimizers crucially depend on the stability of gradient directions across iterations. In high-dimensional settings, ZO gradients tend to exhibit weaker directional correlation~\citep{ghadimi2013stochastic}, so momentum accumulation fails to yield meaningful acceleration. Together, these factors explain why Adam does not consistently outperform SGD in ZO LLM fine-tuning~\citep{zhang2024revisiting,guo2025zeroth,shu2025refining}, even though it can be more effective in FO  training~\citep{glentis2026revisiting}.

\noindent\textbf{Contributions.} Our main contributions are as follows. Firstly, we propose SubZero+, which features two complementary designs in  random subspaces. SubZero+ employs multi-query ZO gradient estimation with the forward-difference scheme, which reduces contamination in the moment estimates without maintaining full-parameter optimizer states. Then it applies Adam directly to the resulting low-dimensional coefficient matrices. SubZero+ stores only two low-dimensional moment matrices per layer and resets them whenever the random bases are refreshed periodically. The former design enhances gradient accumulation by reducing the noise of the gradient estimator, thereby improving the quality of the moment estimates. The latter design performs momentum accumulation in the low-dimensional space, enabling Adam to produce stable and efficient updates while keeping the optimizer state lightweight. With the dual low-dimensionality, SubZero+ achieves both lower memory consumption and improved optimization performance. 

Secondly, we characterize the query-induced noise that enters SubZero+'s coefficient-space moments in the exact-directional limit. Theorem~\ref{thm:first_moment} shows that the $K$-query coefficient estimator is conditionally unbiased and that its covariance and mean-squared error decrease exactly as $1/K$. Theorem~\ref{thm:second_moment} further shows that each coordinate's expected squared write contains an additive global-energy term suppressed by exactly $1/K$. These conditional, single-step identities clarify why multiple queries are especially relevant to the inputs of Adam's accumulators.

Finally, we evaluate SubZero+ on SuperGLUE across model sizes from 1.3B to 32B under both full-parameter tuning and LoRA~\citep{hu2021lora}. Across all settings, SubZero+ consistently outperforms the evaluated ZO baselines, narrowing the performance gap to FO methods while introducing minimal additional memory overhead. These results indicate that multi-query estimation and compact moments in random low-rank subspaces provide a practical pathway for adaptive ZO LLM fine-tuning.

\begin{figure*}[t]
	\centering
	\setlength{\tabcolsep}{3pt}
	\renewcommand{\arraystretch}{1.05}
	\begin{tabular}{ccc}
		\vspace{-0.5em}
		\includegraphics[width=0.31\linewidth]{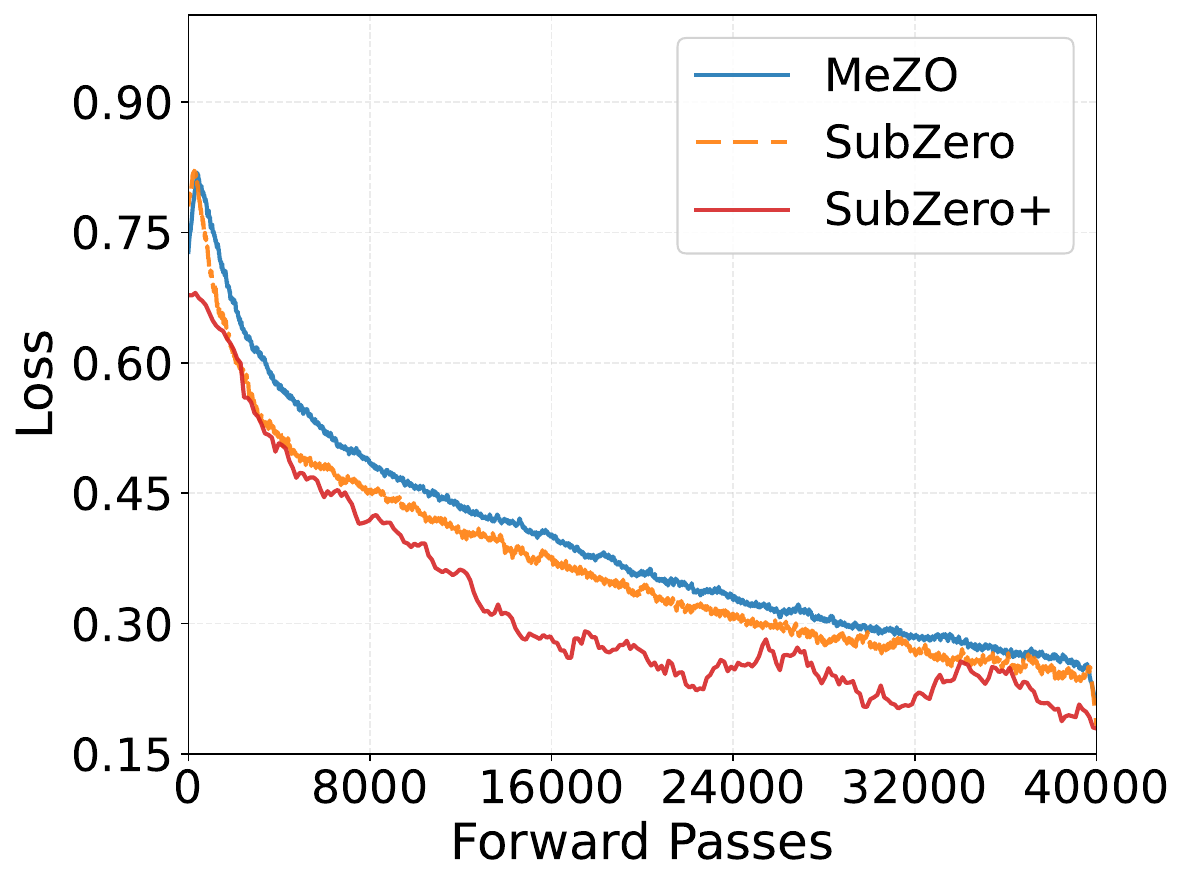} &
		\includegraphics[width=0.31\linewidth]{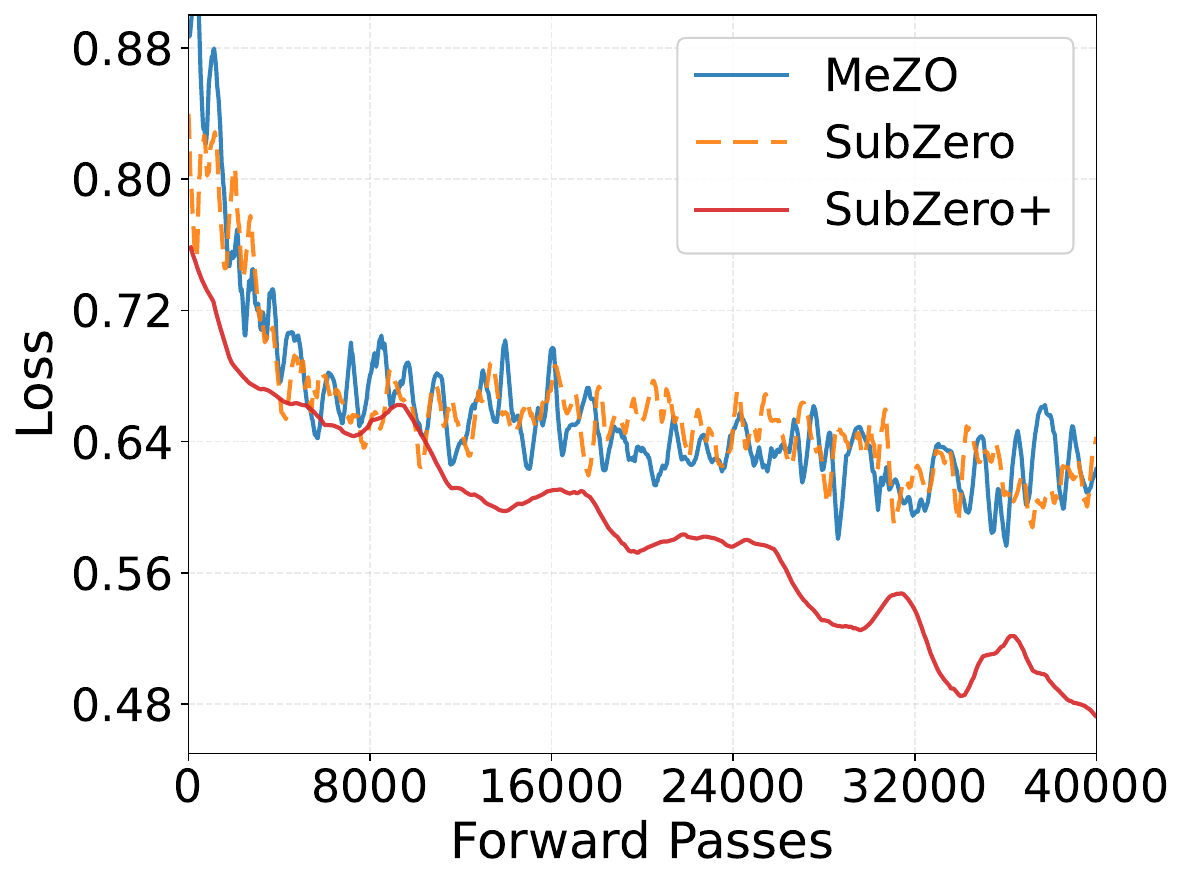} &
		\includegraphics[width=0.31\linewidth]{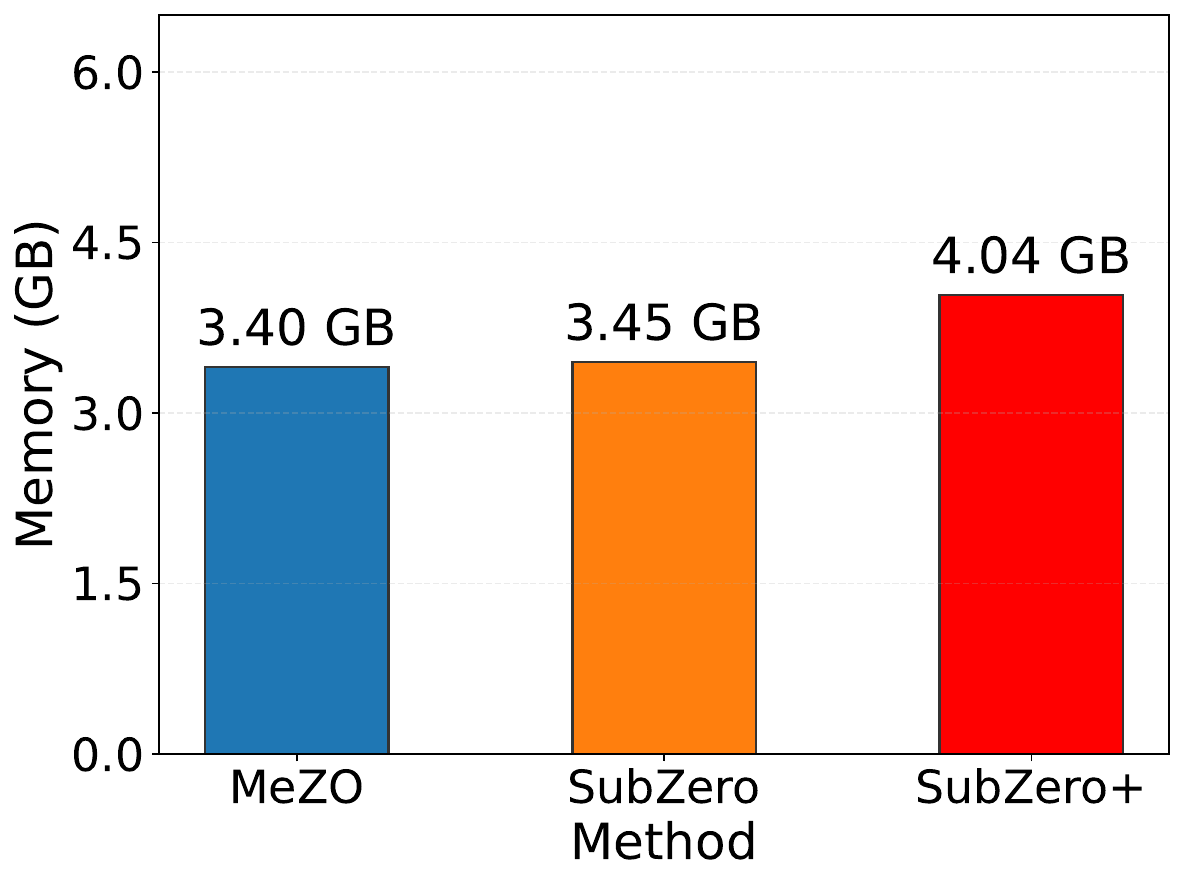} \\
		\footnotesize{(a) OPT-1.3B FT on SST-2} &
		\footnotesize{(b) LLaMA3.1-8B LoRA on WiC} &
		\footnotesize{(c) Memory Cost of (a)} \\
	\end{tabular}
	\vspace{-0.5em}
	\caption{Training dynamics and memory cost of SubZero+. (a) Under full-parameter tuning of OPT-1.3B on SST-2, SubZero+ reaches a lower loss than MeZO and SubZero within the same forward-pass budget.
		(b) Under LoRA tuning of LLaMA3.1-8B on WiC, SubZero+ gives a smoother trajectory and reaches a lower loss.
		(c) For the setting of (a), SubZero+ introduces only $\sim$0.6GB additional GPU memory overhead compared with MeZO and SubZero.}
	\label{fig:LRSensitivity}
\end{figure*}

\section{Related Work}
\label{sec: related_work}
\vspace*{-0.5em}

\noindent\textbf{ZO Gradient Estimation.}
ZO optimization replaces BP with finite-difference gradient estimates that rely only on forward passes. MeZO~\citep{malladi2023fine} pioneered SPSA-based~\citep{spall1992multivariate} ZO fine-tuning for LLMs, achieving inference-level memory usage while exhibiting gradient estimation variance that scales with the parameter dimension. Subsequent work reduces this variance through sparse perturbations~\citep{liu2024sparse,guo2025zeroth} and low-rank subspace projections~\citep{chen2024enhancing,yu2025zeroth}. LOZO~\citep{chen2024enhancing} factorizes perturbations as low-rank matrices, while SubZero~\citep{yu2025zeroth} enforces column-orthonormal projection matrices via QR decomposition. All these methods use a single query per iteration, which limits the achievable variance reduction.

\noindent\textbf{ZO Parameter Update.}
Many ZO methods, including MeZO \citep{malladi2023fine} and SubZero \citep{yu2025zeroth}, apply vanilla SGD to the estimated gradients. LOZO \citep{chen2024enhancing} performs SGDM within a low-rank subspace and requires negligible additional memory. TeZO \citep{sun2025tezo} further exploits temporal low-rank structure across gradients and can be extended to an Adam variant. However, previous attempts at adaptive ZO optimization have consistently failed to outperform SGD in practice~\citep{malladi2023fine,yu2025zeroth,guo2025zeroth}.

In contrast, adaptive ZO optimization extends Adam/RMSProp-style adaptive normalization to the ZO setting. ZO-AdaMM \citep{chen2019zo} generalizes the AdaMM framework by combining momentum-based first-moment accumulation with AMSGrad-style second-moment normalization. ZO-AdaMU \citep{jiang2024zo} relocates momentum into the simulated perturbation step in stochastic gradient approximation.  R-AdaZO \citep{shu2025refining} refines the second-moment update by using the squared first moment.

\noindent\textbf{Multi-Query ZO.}
Averaging multiple independent perturbation estimates can reduce per-step gradient variance~\citep{duchi2015optimal}. However, \citet{lin2025multi} identified the multi-query paradox. Under a fixed query budget, the variance reduction from using more queries per update may be offset by the reduced number of parameter updates. Recent work addresses this trade-off  through layer-wise independent perturbations~\citep{peng2025muzo}, gradient normalization~\citep{dang2025fzoo}, projection alignment~\citep{lin2025multi}, and gradient orthogonalization~\citep{lang2026powering}. While these works address the multi-query paradox through various strategies, they primarily operate in full-parameter space, limiting variance reduction gains.

\noindent\textbf{Theory of Adaptive ZO Update.}
ZO-AdaMM~\citep{chen2019zo} establishes convergence for an adaptive ZO optimizer with a persistent maximum second-moment state. MUZO~\citep{peng2025muzo} leverages multi-query gradient averaging with Adam in full-parameter space, while R-AdaZO \citep{shu2025refining} introduces the variance-aware convergence framework by exploiting the variance reduction effect of first moment estimates and refining second moment estimates accordingly. These advances reveal that adaptively coupling noisy gradient estimates to their current normalizers requires careful theoretical treatment~\citep{taniguchi2024adopt}.  

\section{Preliminaries}
\label{sec: preliminaries}
\vspace*{-0.5em}

We first introduce the foundational SubZero framework that SubZero+ builds upon. We use bold lowercase letters \(\vw\) to denote vectors, bold uppercase letters \(\mW\) to denote matrices, and calligraphic letters \(\mathcal{W}\) to denote sets of matrices. \(\text{vec}(\boldsymbol{W})\) vectorizes \(\boldsymbol{W}\) by stacking its columns into a single column vector, \(\boldsymbol{A}\odot \boldsymbol{B}\)  denotes element-wise multiplication, \(\boldsymbol{A}\otimes \boldsymbol{B}\) denotes the Kronecker product, \(\mathbb{E}[\vx]\) and \(\Var[\vx]\) denote the expectation and variance of the random variable \(\vx\), \(\|\vx\|=\sqrt{\sum_{i=1}^{n}\vx_i^2}\) denotes the \(\ell_2\)-norm, and \(\mathrm{bdiag}(\bm{A}_1,\ldots,\bm{A}_\ell)\) denotes a block diagonal matrix with diagonal blocks \(\bm{A}_1,\ldots,\bm{A}_\ell\).

\noindent\textbf{Problem Setup.}
Consider fine-tuning a pre-trained LLM with $l$ layers. Let the trainable parameters of layer $i$ be represented by a matrix $\mW_i \!\in\! \mathbb{R}^{m_i \times n_i}$ (for example, the weight matrices of linear projections in attention and feed-forward blocks). Let $\mathcal{W}=\{\mW_i\}_{i=1}^{l}$ denote the set of all trainable parameter matrices, and define the flattened parameter vector
$\vw=\big(\mathrm{vec}(\mW_1)^\top,\dots,\mathrm{vec}(\mW_l)^\top\big)^\top \!\in\! \mathbb{R}^d,
$ where $d=\sum_{i=1}^{l} m_i n_i$ is the total number of represented matrix parameters. The fine-tuning objective is to minimize a loss function $\mathcal{L}(\mathcal{W})$ over a training dataset $\mathcal{D}$.

\noindent\textbf{SubZero.}
SubZero~\citep{yu2025zeroth} reduces the variance of gradient estimation by restricting perturbations to a low-rank subspace constructed separately for each layer. For layer $i$, SubZero generates two column-orthonormal projection matrices $\mU_i\in\mathbb{R}^{m_i\times r}$ and $\mV_i\in\mathbb{R}^{n_i\times r}$ by taking the QR decomposition of random Gaussian matrices, where $r \ll \min\{m_i,n_i\}$ is the subspace rank. The layer-wise perturbation is
\begin{align}
	\tilde{\mZ}_i = \mU_i \mZ_i \mV_i^\top \in \mathbb{R}^{m_i \times n_i},
	\label{eq:subzero_pert}
\end{align}
where $\mZ_i\in\mathbb{R}^{r\times r}$ is a low-dimensional random matrix with i.i.d.\ $\mathcal{N}(0,1)$ entries. The loss difference is computed globally as
\begin{align}
	\rho = \frac{\mathcal{L}(\mathcal{W} + \varepsilon \tilde{\mathcal{Z}}; \mathcal{B}) - \mathcal{L}(\mathcal{W} - \varepsilon \tilde{\mathcal{Z}}; \mathcal{B})}{2\varepsilon},
	\label{eq:subzero_rho}
\end{align}
where $\tilde{\mathcal{Z}}=\{\tilde{\mZ}_i\}_{i=1}^{l}$ denotes the set of layer-wise perturbations. The gradient estimate for layer $i$ is 
$\widehat{\nabla}_{\mW_i}\mathcal{L}=\rho\,\tilde{\mZ}_i$,
and parameters are updated using SGD via
$\mW_i \leftarrow \mW_i - \eta\,\widehat{\nabla}_{\mW_i}\mathcal{L}$, where $\eta$ denotes the learning rate.

\section{The Proposed SubZero+}
\label{sec: method}
\vspace*{-0.5em} 

SubZero+ introduces two design choices to make adaptive ZO optimization practical in random subspaces. First, it employs multi-query forward differences to reduce direction-sampling noise in the coefficient estimate. Second, it runs Adam entirely in the resulting low-dimensional coefficient matrices, so its first- and second-moment buffers remain negligible relative to the model parameters.

\subsection{Multi-Query Estimation with Forward-Difference}
\label{sec:multi_query}

SubZero estimates the gradient from a single perturbation. To improve the coefficient signal used by Adam, SubZero+ averages multiple independent perturbation queries. Theorem~\ref{thm:first_moment} separates the practical one-sided estimator's finite-difference bias from its query variance, while Theorem~\ref{thm:second_moment} shows that averaging reduces additive second-moment inflation by $1/K$ in the exact-directional limit. Rather than computing full-parameter gradient estimates explicitly, we accumulate multi-query estimates directly within the low-dimensional coefficient space. While more queries reduce per-update direction-sampling variance, they also reduce the total number of updates available under a fixed forward-evaluation budget. The optimal query count is therefore problem-dependent and is explored in Subsection~\ref{sec:ablation}. We employ forward differences rather than central differences, requiring $K$ perturbed evaluations plus one shared baseline evaluation per update.

Formally, let \(\mU_i \in \mathbb{R}^{m_i \times r}\) and \(\mV_i \in \mathbb{R}^{n_i \times r}\) be fixed projection matrices for layer \(i\) at the current step. 
They are two column-orthonormal projection matrices generated by QR decomposition of random Gaussian matrices.
For query \(k = 1,\dots,K\), we independently sample \(\mZ_i^{(k)} \in \mathbb{R}^{r \times r}\) with i.i.d.\ \(\mathcal{N}(0,1)\) entries and construct the full-parameter perturbation
\begin{align}
	\tilde{\mZ}_i^{(k)} = \mU_i \mZ_i^{(k)} \mV_i^\top .
	\label{eq:mq_pert}
\end{align}
Let \(\tilde{\mathcal{Z}}^{(k)} = \{\tilde{\mZ}_i^{(k)}\}_{i=1}^{l}\) denote the collection of perturbations across all \(l\) layers. 
We compute the baseline loss \(\ell_0 = \mathcal{L}(\mathcal{W}; \mathcal{B})\) and the perturbed loss for each query
\begin{align}
	\ell_k = \mathcal{L}(\mathcal{W} + \varepsilon \tilde{\mathcal{Z}}^{(k)}; \mathcal{B}) .
	\label{eq:mq_loss}
\end{align}
Employing a forward-difference scheme, the multi-query gradient estimate for layer \(i\) becomes
\begin{align}
	\widehat{\nabla}_{\mW_i} \mathcal{L}(\mathcal{W}; \mathcal{B})
	= \frac{1}{K} \sum_{k=1}^{K} \frac{\ell_k - \ell_0}{\varepsilon}\, \tilde{\mZ}_i^{(k)} 
	= \mU_i \left(\frac{1}{K} \sum_{k=1}^{K} \frac{\ell_k - \ell_0}{\varepsilon}\, \mZ_i^{(k)} \right)\mV_i^\top 
	= \mU_i \overline{\mZ}_i \mV_i^\top .
	\label{eq:mq_grad_est}
\end{align}

\begin{figure}[t]
	\begin{minipage}[t]{.45\textwidth}
		\begin{algorithm}[H]
			\caption{SignCorrectedQR$(r, m_i, n_i)$}
			\label{alg:haar_qr}
			\begin{algorithmic}[1]
				\REQUIRE{Rank $r$, dimensions $m_i$, $n_i$.}
				\STATE Draw $\boldsymbol{\Omega}_U \in \mathbb{R}^{m_i \times r}$, $\boldsymbol{\Omega}_V \in \mathbb{R}^{n_i \times r}$ with i.i.d.\ $\mathcal{N}(0,1)$
				\STATE $\tilde{\mU}_i, \mR_U \leftarrow$ QR$(\boldsymbol{\Omega}_U)$; $\tilde{\mV}_i, \mR_V \leftarrow$ QR$(\boldsymbol{\Omega}_V)$
				\STATE $\vs_U \leftarrow (\mathrm{sgn}(\mR_{U,11}), \ldots, \mathrm{sgn}(\mR_{U,rr}))^\top$; $\vs_V \leftarrow (\mathrm{sgn}(\mR_{V,11}), \ldots, \mathrm{sgn}(\mR_{V,rr}))^\top$
				\STATE $\mU_i \leftarrow \tilde{\mU}_i \, \mathrm{diag}(\vs_U)$; $\mV_i \leftarrow \tilde{\mV}_i \, \mathrm{diag}(\vs_V)$
				\RETURN $\mU_i$, $\mV_i$
			\end{algorithmic}
		\end{algorithm}
	\end{minipage}
	\hfill
	\begin{minipage}[t]{.54\textwidth}
		\begin{algorithm}[H]
			\caption{PerturbParams$(\mathcal{W}, \mathcal{U}, \mathcal{V}, r, \varepsilon, s)$}
			\label{alg:subspaceperturb}
			\begin{algorithmic}[1]
				\REQUIRE{Model parameters $\mathcal{W}$, projection sets $\mathcal{U}$, $\mathcal{V}$, rank $r$, perturbation scale $\varepsilon$, seed $s$.}
				\vspace{0.53em}
				\STATE Reset RNG with seed $s$
				\STATE $\mathcal{Z} \leftarrow \emptyset$
				\FOR{$i = 1, 2, \dots, l$}
				\STATE Generate $\boldsymbol{Z}_i \in \mathbb{R}^{r \times r}$ with i.i.d.\ $\mathcal{N}(0,1)$ entries
				\STATE $\mathcal{Z} \leftarrow \mathcal{Z} \cup \{\boldsymbol{Z}_i\}$
				\STATE $\boldsymbol{W}_i \leftarrow \boldsymbol{W}_i + \varepsilon \, \boldsymbol{U}_i \boldsymbol{Z}_i \boldsymbol{V}_i^\mathsf{T}$
				\ENDFOR
				\RETURN $\mathcal{W}$, $\mathcal{Z}$
			\end{algorithmic}
		\end{algorithm}
	\end{minipage}
	\vspace{-0.7em}
	\begin{algorithm}[H]
		\caption{SubZero+}
		\label{alg:ss_mezo}
		\begin{algorithmic}[1]
			\REQUIRE{Parameter matrices $\boldsymbol{W}_i \!\in\! \mathbb{R}^{m_i \times n_i}$ for $i\! =\! 1, \dots, l$, loss $\mathcal{L}$, step budget $T$, perturbation scale $\varepsilon$, learning rate $\eta$, subspace change frequency $F$, rank $r$, perturbation-query count $K$, Adam parameters $\beta_1,\beta_2,\epsilon_{\mathrm{adam}}$.}
			\FOR{$t = 1, 2, \dots, T$}
			\STATE Sample a minibatch $\mathcal{B}^t \subset \mathcal{D}$
			\FOR{$i = 1, 2, \dots, l$}
			\IF{$(t\!-\!1) \bmod F \equiv 0$}
			\STATE $\boldsymbol{U}_{i}^{t}, \boldsymbol{V}_{i}^{t} \leftarrow$ SignCorrectedQR$(r, m_i, n_i)$
			\STATE $\mM_i^{t-1} \leftarrow \boldsymbol{0}$; $\mS_i^{t-1} \leftarrow \boldsymbol{0}$   \hfill \COMMENT{restart Adam states before recursion}
			\STATE $c_t \leftarrow 1$ \hfill \COMMENT{restart bias-correction counter}
			\ELSE
			\STATE $\boldsymbol{U}_{i}^{t} \leftarrow \boldsymbol{U}_{i}^{t-1}$; $\boldsymbol{V}_{i}^{t} \leftarrow \boldsymbol{V}_{i}^{t-1}$
			\STATE $c_t \leftarrow c_{t-1} + 1$
			\ENDIF
			\ENDFOR
			
			\STATE Compute baseline loss $\ell_0^t \leftarrow \mathcal{L}(\mathcal{W}^t; \mathcal{B}^t)$
			\STATE Initialize coefficient gradients $\overline{\mZ}_i^{t} \leftarrow \boldsymbol{0} \in \mathbb{R}^{r\times r}$ for all $i$
			
			\FOR{$k = 1, 2, \dots, K$}
			\STATE Sample a random seed $s_k^t$
			\STATE $\mathcal{W}^t, \mathcal{Z}_k^t \leftarrow$ PerturbParams$(\mathcal{W}^t, \mathcal{U}^t, \mathcal{V}^t, r, \varepsilon, s_k^t)$
			\STATE $\ell_k^t \leftarrow \mathcal{L}(\mathcal{W}^t; \mathcal{B}^t)$
			\STATE $\rho_k^t \leftarrow (\ell_k^t - \ell_0^t) / \varepsilon$
			\STATE $\mathcal{W}^t, \_ \leftarrow$ PerturbParams$(\mathcal{W}^t, \mathcal{U}^t, \mathcal{V}^t, r, -\varepsilon, s_k^t)$
			\FOR{$i = 1, 2, \dots, l$}
			\STATE $\overline{\mZ}_i^t \leftarrow \overline{\mZ}_i^t + \rho_k^t~ \boldsymbol{Z}_{k,i}^t / K$
			\ENDFOR
			\ENDFOR

			\FOR{$i = 1, 2, \dots, l$}
			
			\STATE Compute $\mM_i^{t}$ and $\mS_i^{t}$ by Eqs.~(\ref{eq:adam_m1}) and (\ref{eq:adam_m2})
			\STATE Bias-correct the two moments with local counter $c_t$ by Eq.~(\ref{eq:adam_bias})
			\STATE Compute $\boldsymbol{W}_i^{t+1}$ by Eq.~(\ref{eq:adam_update})
			\ENDFOR
			\ENDFOR
			\RETURN $\mathcal{W}^{T+1}$
		\end{algorithmic}
	\end{algorithm}
	\vspace{-1cm}
\end{figure}

\subsection{Low-Rank Subspace Adam}
\label{sec:subspace_adam}
SubZero applies SGD to all coefficient directions with a common learning rate. In contrast, coefficient scales can differ substantially across layers and coordinates. SubZero+ therefore runs Adam entirely in the \(r\times r\) coefficient space. For layer \(i\) at step \(t\), let \(\overline{\mZ}_i^t\) denote the estimate in Eq.~(\ref{eq:mq_grad_est}). We maintain two \(r\times r\) buffers
\begin{align}
	\mM_i^t &= \beta_1 \mM_i^{t-1} + (1-\beta_1)\overline{\mZ}_i^t, \label{eq:adam_m1}\\
	\mS_i^t &= \beta_2 \mS_i^{t-1} + (1-\beta_2)(\overline{\mZ}_i^t\odot\overline{\mZ}_i^t), \label{eq:adam_m2}
\end{align}
where $0\le\beta_1,\beta_2<1$, and $\epsilon_{\mathrm{adam}}>0$.  Since the projection bases $\mU_i^t$ and $\mV_i^t$ remain fixed throughout 
each $F$-step block, the moment buffers preserve a consistent coordinate system across iterations within the block. To correct for EMA initialization at block 
entrances, we apply local bias correction using the within-
block counter $c_t = 1 + ((t-1) \bmod F)$ rather than the 
global iteration count. The locally bias-corrected states are
\begin{align}
	\widehat{\mM}_i^t=\mM_i^t/(1-\beta_1^{c_t}),\qquad
	\widehat{\mS}_i^t=\mS_i^t/(1-\beta_2^{c_t}).
	\label{eq:adam_bias}
\end{align}
The ambient update is
\begin{align}
	\mW_i^{t+1}
	= \mW_i^t - \eta\, \mU_i^t
	\left(\widehat{\mM}_i^t\Big/\left(\sqrt{\widehat{\mS}_i^t}+\epsilon_{\mathrm{adam}}\right)\right)
	(\mV_i^t)^\top. \label{eq:adam_update}
\end{align}
All square roots and divisions in this update are element-wise. The projection bases are periodically refreshed every \(F>1\) steps as in SubZero. The moment buffers cannot be transferred across basis refreshes because successive random subspaces are nearly orthogonal in low-rank settings, rendering previous coordinate statistics incompatible. Therefore, before the recursion at a refresh step, we set the pre recursion states \(\mM_i^{t-1}\) and \(\mS_i^{t-1}\) to zero and restart \(c_t\). While first-moment transport across basis refreshes is theoretically possible, preliminary experiments show negligible performance gains while complicating implementation. We therefore adopt a simpler reset strategy and leave transport mechanisms for future work. Each matrix layer contributes only \(2r^2\) moment floats. Separately handled one-dimensional parameters have their own scalar Adam states.

\subsection{Final Algorithm}
\label{sec:implementation}

Algorithm~\ref{alg:haar_qr} constructs the sign-corrected projection matrices following the standard QR-based approach to Haar sampling~\citep{mezzadri2007generate}. Note that standard QR implementations (e.g., \texttt{torch.linalg.qr}) do not enforce a positive diagonal in the triangular factor. Therefore, we apply a sign correction for unique decomposition.  Algorithm~\ref{alg:subspaceperturb} generates and applies the perturbations, and Algorithm~\ref{alg:ss_mezo} presents the complete SubZero+ training loop. Let \(K\) be the number of perturbation queries per update: each iteration evaluates \(K\) perturbed losses plus one additional baseline loss, for a total of \(K+1\) forward passes.  SubZero+ remains entirely BP-free and inherits SubZero’s compatibility with full-parameter tuning (FT), LoRA~\citep{hu2021lora}, prefix tuning~\citep{li2021prefix}, and prompt tuning~\citep{lester2021power}. For LoRA, following SubZero, we reshape each trainable two-dimensional factor into an approximately square matrix with the same number of elements before constructing the random subspace, and reshape each perturbation and update back to its original shape.

\section{Theoretical Analysis}
\label{sec:theory}
\vspace*{-0.5em}

We analyze how multi-query averaging affects both the coefficient estimate and its squared value written to Adam. The results are conditional, single-step statements for a fixed optimization state and minibatch.

Let $q=lr^2$ and $\mP=\operatorname{blkdiag}_{i=1}^l(\mV_i\otimes\mU_i)\in\mathbb R^{d\times q}$. For query $k$, stack the coefficient directions as $\vz_k=(\mathrm{vec}(\mZ_1^{(k)})^\top,\ldots,\mathrm{vec}(\mZ_l^{(k)})^\top)^\top$. Since $\mathrm{vec}(\mU_i\mZ_i^{(k)}\mV_i^\top)=(\mV_i\otimes\mU_i)\mathrm{vec}(\mZ_i^{(k)})$, $\mP\vz_k$ is exactly the concatenated perturbation in Eq.~(\ref{eq:mq_pert}), and $\mP^\top\mP=\mI_q$. For minibatch $\mathcal B$, let $f_{\mathcal B}(\vw)=\mathcal L(\mathcal W;\mathcal B)$, $\va=\mP^\top\nabla_{\vw}f_{\mathcal B}(\vw)$, and $a_p=\boldsymbol e_p^\top\va$ for $p\in[q]:=\{1,\ldots,q\}$, where $\boldsymbol e_p$ is the $p$th standard basis vector.

We assume that all calls evaluate the same minibatch-loss function, with non-query evaluation randomness fixed and shared. Let $\mathcal G$ be the sigma-field generated by the pre-query history, $\mathcal B$, $\vw$, $\mP$, and this shared randomness; hence $f_{\mathcal B}$ is fixed given $\mathcal G$. Conditionally on $\mathcal G$, draw $\vz_1,\ldots,\vz_K\stackrel{\mathrm{i.i.d.}}{\sim}\mathcal N(\vzero,\mI_q)$. When $f_{\mathcal B}$ is differentiable at $\vw$, the practical estimator and its pointwise exact-directional limit are
\begin{align}
\widehat\va_\varepsilon
&=\frac1K\sum_{k=1}^K
\frac{f_{\mathcal B}(\vw+\varepsilon\mP\vz_k)-f_{\mathcal B}(\vw)}{\varepsilon}\vz_k,
&
\widehat\va
&:=\lim_{\varepsilon\downarrow0}\widehat\va_\varepsilon
=\frac1K\sum_{k=1}^K\vz_k\vz_k^\top\va,
\label{eq:fd_and_directional_estimators}
\end{align}
where the limit fixes the sampled directions, $K\in\mathbb N_{+}$, and $\varepsilon>0$. Its $i$th block is $\mathrm{vec}(\overline{\mZ}_i)$ from Eq.~(\ref{eq:mq_grad_est}).

\begin{theorem}[Exact-directional coefficient estimation]
\label{thm:first_moment}
\begin{align}
\mathbb E\!\left[\widehat\va\mid\mathcal G\right]&=\va,
&
\operatorname{Cov}(\widehat\va\mid\mathcal G)
&=\frac1K\left(\va\va^\top+\|\va\|^2\mI_q\right),
&
\mathbb E\!\left[\|\widehat\va-\va\|^2\mid\mathcal G\right]
&=\frac{q+1}{K}\|\va\|^2.
\label{eq:exact_directional_error}
\end{align}
\end{theorem}
Averaging $K$ queries leaves the coefficient estimate unbiased while reducing its covariance and mean-squared error by $1/K$. For $\va\ne\vzero$, the relative MSE is $(q+1)/K$, depending on coefficient dimension $q$ rather than represented dimension $d$.

\begin{theorem}[Exact-directional squared-write inflation]
\label{thm:second_moment}
For $\vx\in\mathbb R^q$, let $\vx^{\odot2}=\vx\odot\vx$ denote coordinatewise squaring and let $\boldsymbol 1_q$ be the all-ones vector. For every $p\in[q]$, define $\widehat a_p=\boldsymbol e_p^\top\widehat\va$. Then
\begin{align}
\mathbb E\!\left[\widehat\va^{\odot 2}\mid\mathcal G\right]
&=\left(1+\frac1K\right)\va^{\odot 2}
+\frac{\|\va\|^2}{K}\boldsymbol{1}_q,
\label{eq:second_moment_identity}\\
\mathbb E\!\left[\widehat a_p^2\mid\mathcal G\right]
&=\left(1+\frac1K\right)a_p^2+\frac{\|\va\|^2}{K}.
\label{eq:coordinate_second_moment}
\end{align}
\end{theorem}
The global-energy term $\|\va\|^2/K$ means that even if $a_p=0$, coordinate $p$ receives positive expected second-moment input when $\va\ne\vzero$; averaging $K$ queries suppresses it by $1/K$.

\noindent\textbf{Connection to Existing Convergence Theory.}
SubZero and LOZO establish convergence for related random-subspace ZO updates under standard smoothness assumptions~\citep{yu2025zeroth,chen2024enhancing}. Rather than repeat this generic analysis, Theorems~\ref{thm:first_moment} and~\ref{thm:second_moment} isolate SubZero+'s new multi-query estimation error and its effect on adaptive moments. Complete proofs are in Appendix~\ref{appendix:theory}.

\section{Experiments}
\label{sec:experiment}
\vspace*{-0.5em} 

\begin{table*}[t]
\vspace{-5pt}
\caption{
    Main results on autoregressive LLMs under the FT scheme. SubZero+ uses $K=99$ perturbation queries plus one baseline evaluation per update. All methods use a matched 40K total forward-evaluation budget.
}
\vspace{1.5pt}
\centering
\setlength{\tabcolsep}{3pt}
    \resizebox{\textwidth}{!}{\begin{tabular}{lccccccccccc}
    \toprule
    Model & Method & \multicolumn{1}{c}{SST-2} & \multicolumn{1}{c}{ RTE} & \multicolumn{1}{c}{ BoolQ} & \multicolumn{1}{c}{ WIC} & \multicolumn{1}{c}{ MultiRC} & \multicolumn{1}{c}{COPA} & \multicolumn{1}{c}{ReCoRD} & \multicolumn{1}{c}{SQuAD} & \multicolumn{1}{c}{DROP} & \multicolumn{1}{c}{Avg.}\\
    & Task type & \multicolumn{5}{c}{------ Classification ------} & \multicolumn{2}{c}{--- Multi-choice ---} & \multicolumn{2}{c}{--- Generation ---} \\
    \midrule
    \multirow{9}{*}{OPT-1.3B} & Zero-shot & 53.5 & 53.4 & 45.5 & 57.5 & 45.4 & 75.0 & 70.5 & 27.2 & 11.1 & 48.8\\
    & FO (Adam) & 92.0 & 78.0 & 75.8 & 66.0 & 70.9 & 77.0 & 71.1 & 81.7 & 31.1 & 71.5\\
    \cmidrule(lr){2-12}
    & MeZO & 90.2 & 64.3 & 65.0 & 58.3 & 59.7 & 74.0 & 71.0 & 76.4 & 22.5 & 64.6\\
    & SubZero & 90.9 & \textbf{67.5} & 65.4 & 58.1 & 59.5 & 75.0 & 70.9 & 76.4 & 25.4 & 65.5\\
    & LOZO & 91.1 & 67.2 & 66.7 & 58.9 & 61.4 & 75.0 & 72.1 & 74.5 & \textbf{25.5} & 65.8\\
    & TeZO & 91.3 & 63.2 & 63.4 & 58.5 & \textbf{62.0} & \textbf{77.0} & 71.8 & 74.1 & 22.2 & 64.8\\
    & ZO-Muon & 91.7 & 67.1 & 66.2 & 58.8 & 60.9 & \textbf{77.0} & 72.0 &  76.1 & 24.6 & 66.0\\
    & SubZero+ & \textbf{92.1} & 65.7 & \textbf{66.9} & \textbf{59.4} & 61.4 &  76.0 & \textbf{72.3} & \textbf{77.2} & 25.4 & \textbf{66.3}\\
    \midrule
    \multirow{9}{*}{LLaMA3.1-8B} & Zero-shot & 59.8 & 46.6 & 76.3 & 59.6 & 62.3 & 64.9 & 83.7 & 85.0 & 28.5 & 63.0\\
    & FO (Adam) & 95.9 & 88.1 & 87.7 & 76.3 & 89.3 & 84.0 & 75.8 & 91.4 & 70.1 & 84.3\\
    \cmidrule(lr){2-12}
    & MeZO & 91.6 & \textbf{85.9} & 83.3 & 66.3 & 86.7 & 89.0 &  82.6 & 90.3 & 65.1 & 82.3\\
    & SubZero & 91.2 & 83.8 & 83.3 & 65.5 & 87.2 & 90.0 & 82.6 &  90.9  & \textbf{67.8} & 82.5\\
    & LOZO & 91.2 & 84.5 & 84.5 & 64.3 & \textbf{87.9} & 90.0 & 81.8 & 89.8 & 66.6 & 82.3\\
    & TeZO & 91.2 & 82.7 & \textbf{84.7} & 64.3 & 87.6 & 89.0 & 81.4 & 90.2 & 65.2 & 81.8\\
    & ZO-Muon &  93.7 & 84.8 & 84.6 & 66.9 & 87.7 & 90.0 & \textbf{83.1} & \textbf{91.3} & 67.0 & 83.2\\
    & SubZero+ & \textbf{95.4} & 84.8 & 84.2 &  \textbf{69.6} & 87.2 & \textbf{92.0} & 82.0 & 90.7 & 67.5 & \textbf{83.7}\\   
    \midrule
    \multirow{9}{*}{OPT-13B} & Zero-shot & 58.8 & 59.6 & 59.0 & 55.0 & 46.9 & 80.0 & 81.2 & 46.2 & 14.6 & 55.7\\
    & FO (Adam) & 95.1 & 83.4 & 80.6 & 68.0 & 78.9 & 88.0 & 81.8 & 87.0 & 34.7 & 77.5\\
    \cmidrule(lr){2-12}
    & MeZO & 91.3 & 67.5 & 67.1 & 59.1 & 61.5 & 87.0 & 81.4 & 83.6 &  30.6 & 69.9\\
    & SubZero & 90.8 & 66.4 & 68.3 & 58.3 & 66.8 & 88.0 & 81.2 & 83.6 & 29.2 & 70.3\\
    & LOZO & 92.4 & 67.1 & 68.0 & 57.4 & 62.2 & \textbf{89.0} & 81.0 & 84.2 & 29.6 & 70.1\\
    & TeZO & 92.3 & 72.2 & 68.1 & 62.1 & \textbf{67.0} & 84.0 & 81.7 & 82.7 & 28.2 & 70.9\\
    & ZO-Muon & \textbf{94.4} & 67.5 & 71.9 & 60.5 & 63.3 & \textbf{89.0} & \textbf{81.8} & \textbf{85.7} & \textbf{31.1} & 71.7\\  
    & SubZero+ & 93.5 & \textbf{74.0} & \textbf{75.2} & \textbf{63.2} & 64.7 & \textbf{89.0} & 81.7 & 85.2 & 30.6 & \textbf{73.0}\\
    \bottomrule
    \end{tabular}}
\vspace{-1em}
\label{tab:autoregressive_main}
\end{table*}

\begin{figure}[t]
\begin{minipage}{\textwidth}
	\centering
	\begin{minipage}[b]{0.32\textwidth}
		\centering
		\includegraphics[width=\textwidth]{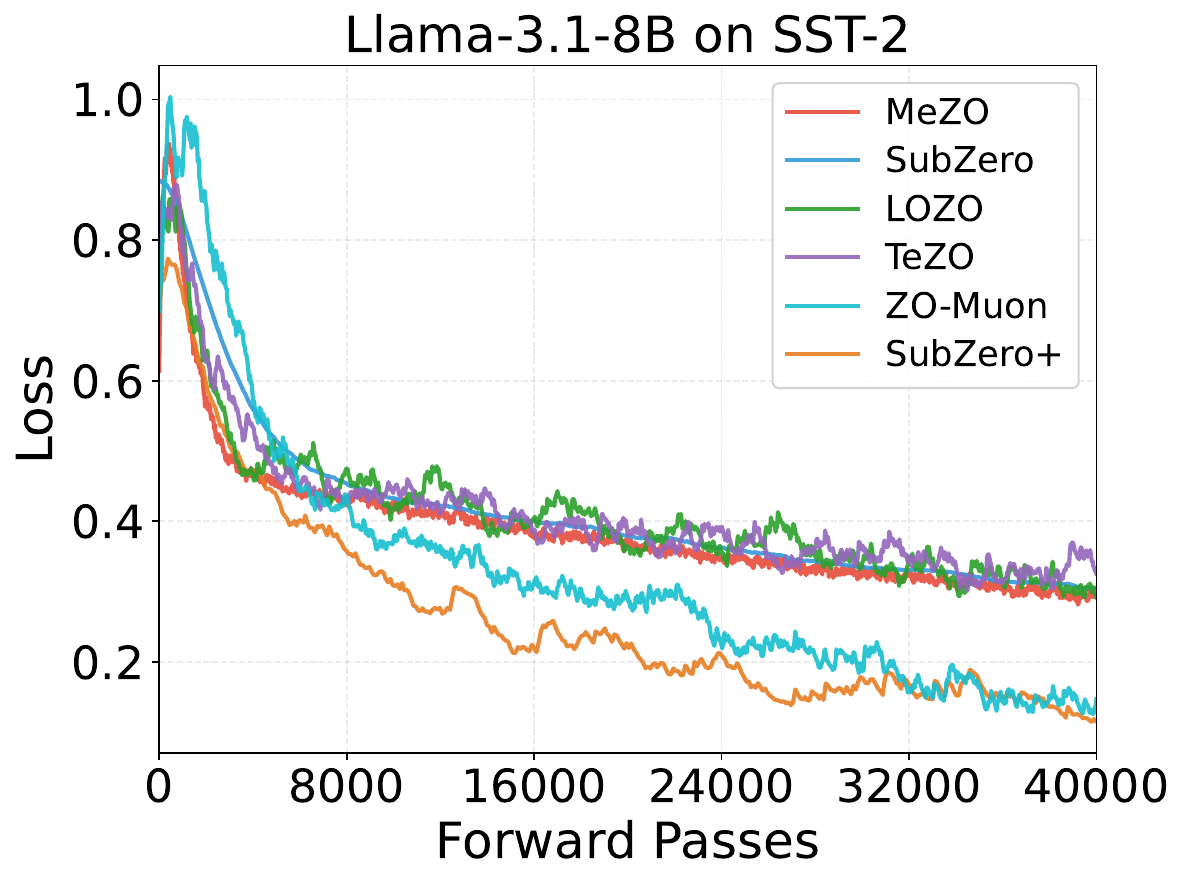}
	\end{minipage}
	\hfill
	\begin{minipage}[b]{0.32\textwidth}
		\centering
		\includegraphics[width=\textwidth]{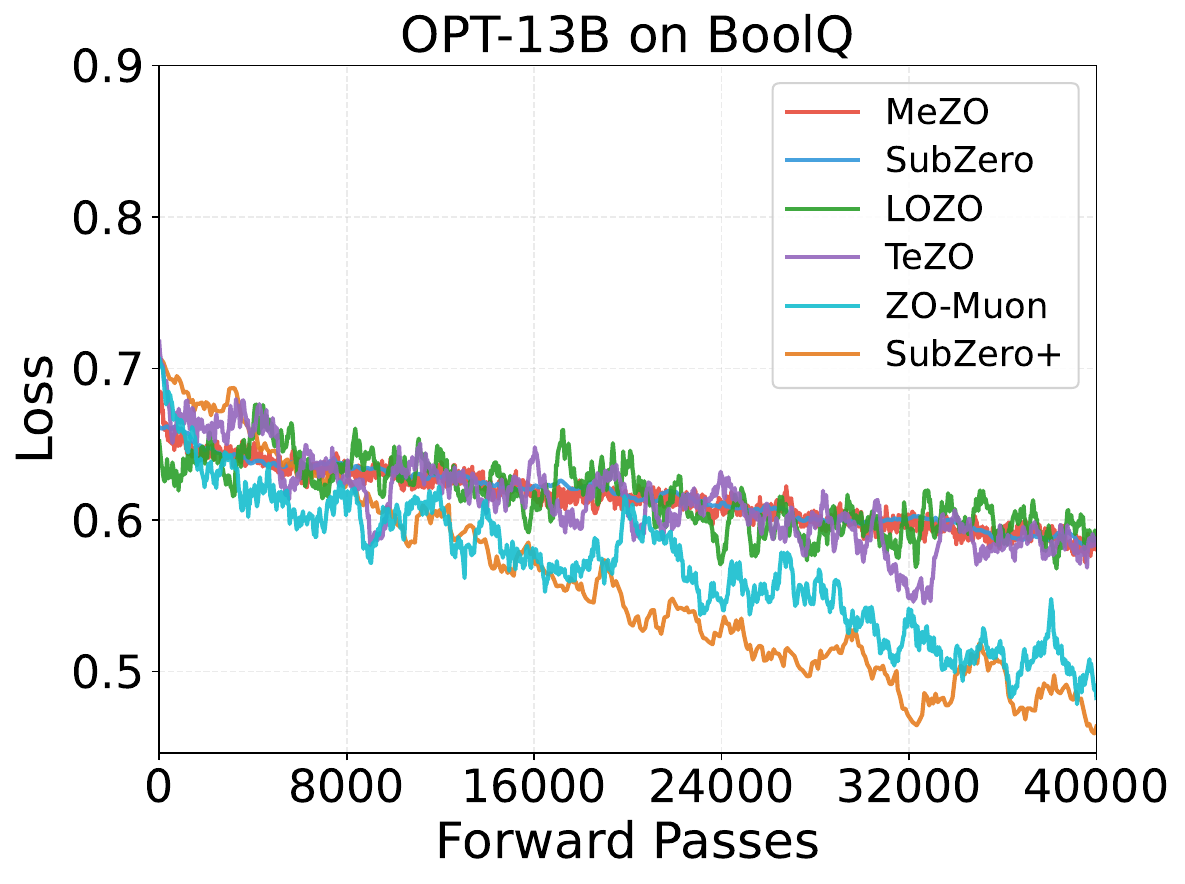}
	\end{minipage}
	\hfill
	\begin{minipage}[b]{0.32\textwidth}
		\centering
		\includegraphics[width=\textwidth]{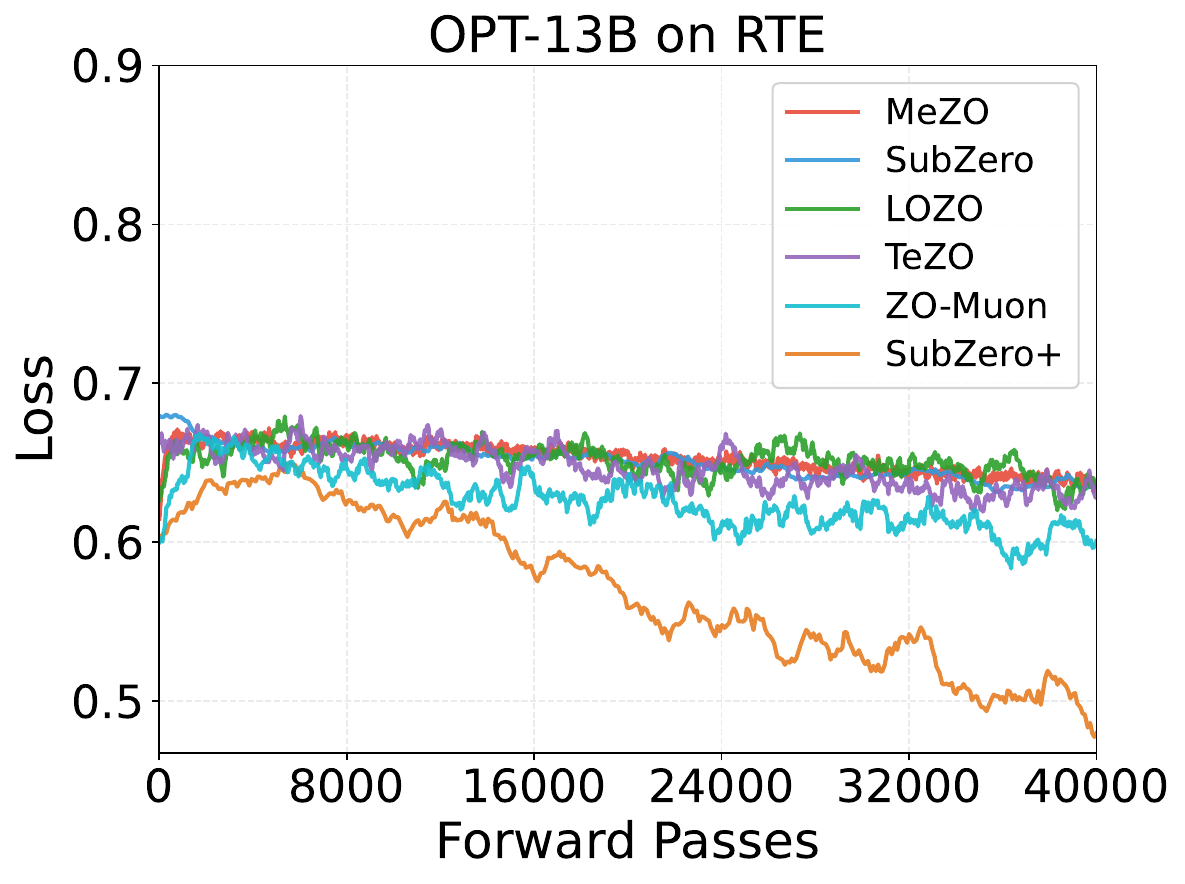}
	\end{minipage}
\end{minipage}
\vspace{-1em}
\caption{Selected loss trajectories corresponding to Table~\ref{tab:autoregressive_main}.}
\label{fig:loss_curves_table1}
\vspace{-1em}
\end{figure}

We conduct comprehensive experiments to evaluate SubZero+ against state-of-the-art ZO baselines across diverse models, tasks, and fine-tuning schemes. Due to space constraints, detailed hyperparameter configurations and additional results are provided in Appendix~\ref{appendix:results} and~\ref{appendix:additional_results}.

\begin{table*}[tb]
\vspace{-15pt}
\caption{
    Main results on autoregressive LLMs under the LoRA scheme. The best result in each model/task column, including Avg., is bolded.}
\vspace{1.5pt}
\renewcommand{\arraystretch}{1}
\centering
\small
\setlength{\tabcolsep}{2pt}
\resizebox{\textwidth}{!}{\begin{tabular}{lccccccccccc}
    \toprule
    Model & Method & \multicolumn{1}{c}{SST-2} & \multicolumn{1}{c}{RTE} & \multicolumn{1}{c}{BoolQ} & \multicolumn{1}{c}{WIC} & \multicolumn{1}{c}{MultiRC} & \multicolumn{1}{c}{COPA} & \multicolumn{1}{c}{ReCoRD} & \multicolumn{1}{c}{SQuAD} & \multicolumn{1}{c}{DROP} & Avg.\\
    & Task type & \multicolumn{5}{c}{------ Classification ------} & \multicolumn{2}{c}{--- Multi-choice ---} & \multicolumn{2}{c}{--- Generation ---} \\
    \midrule
    \multirow{6}{*}{OPT-13B} & MeZO  & \textbf{93.6} & 62.1 & 69.1 & 56.9 & 57.8 & 86.0 & 81.4 & 83.0 & 28.7 & 68.7\\
    & SubZero  & 92.2 & 62.5 & 66.7 & 56.7 & 58.3 & \textbf{87.0} & 81.8 & \textbf{83.7} & 28.3 & 68.6\\
    & LOZO & 92.8 & 61.4 & 64.7 & 55.5 & \textbf{60.9} & 85.0 & \textbf{82.1} & \textbf{83.7} & \textbf{29.5} & 68.4\\
    & TeZO & 92.1 & 65.3 & 66.0 & 59.4 & 59.5 & 85.0 & 81.0 & 80.7 & 29.0 & 68.7\\
    & ZO-Muon & 92.9 & 65.3 & 65.9 & 54.9 & 60.0 & 84.0 & 81.3 & 79.4 & 27.6 & 67.9\\
    & SubZero+  & 93.0 & \textbf{71.5} & \textbf{73.9} & \textbf{60.2} & 60.5 & 86.0 & 80.4 & 82.5 & 28.6 & \textbf{70.7}\\
    \midrule
    \multirow{6}{*}{LLaMA3.1-8B} & MeZO  & 91.3 & 78.0 & 83.6 & 61.0 & 85.8 & \textbf{92.0} & 83.8 & 90.4 & 65.4 & 81.3\\
    & SubZero  & 90.6 & 83.8 & 83.5 & 61.1 & 87.3 & 88.0 & 83.5 & 90.4 & 65.3 & 81.5\\
    & LOZO & 89.2 & 78.0 & 83.3 & 64.9 & 86.6 & 87.0 & 82.2 & 90.1 & 66.8 & 80.9\\
    & TeZO & 87.2 & 79.8 & 82.1 & 61.1 & 86.8 & 86.0 & 82.6 & 90.8 & 66.3 & 80.3\\
    & ZO-Muon & 89.1 & \textbf{86.6} & 83.6 & 63.3 & 85.8 & 88.0 & 81.5 & 90.9 & 65.6 & 81.6\\
    & SubZero+  & \textbf{93.3} & 84.8 & \textbf{84.1} & \textbf{69.0} & \textbf{87.6} & 89.0 & \textbf{84.3} & \textbf{91.1} & \textbf{66.9} & \textbf{83.3}\\
    \midrule
\multirow{6}{*}{OPT-30B} & MeZO  & 92.6 & 68.2 & 63.6 & 60.0 & 65.4 & \textbf{86.0} & 80.5 & 83.6 & 30.6 & 70.1 \\
& SubZero  & 93.0 & 65.3 & 61.4 & 58.2 & 64.0 & 85.0 & \textbf{81.7} & 83.2 & 32.5 & 69.4\\
& LOZO & 93.3 & 66.4 & 64.5 & 60.5 & 60.8 & 85.0 & 81.2 & 82.8 & 32.9 & 69.7\\
& TeZO & 91.5 & 68.2 & 65.5 & \textbf{61.9} & 61.3 & 85.0 & 80.3 & 81.1 & 30.6 & 69.5\\
& ZO-Muon & 90.9 & 65.7 & 65.9 & 56.1 & 60.4 & 83.0 & 80.7 & 78.3 & 29.2 & 67.8\\
& SubZero+  & \textbf{94.2} & \textbf{72.6} & \textbf{71.5} & 60.3 & \textbf{67.0} & \textbf{86.0} & 80.9 & \textbf{84.7} & \textbf{33.0} & \textbf{72.2} \\
    \bottomrule
    \end{tabular}}
\vspace{-10pt}
\label{tab:lora_main}
\end{table*}

\begin{figure*}[t]
	\centering
	\begin{minipage}[t]{0.32\textwidth}
		\centering
		\includegraphics[width=\linewidth]{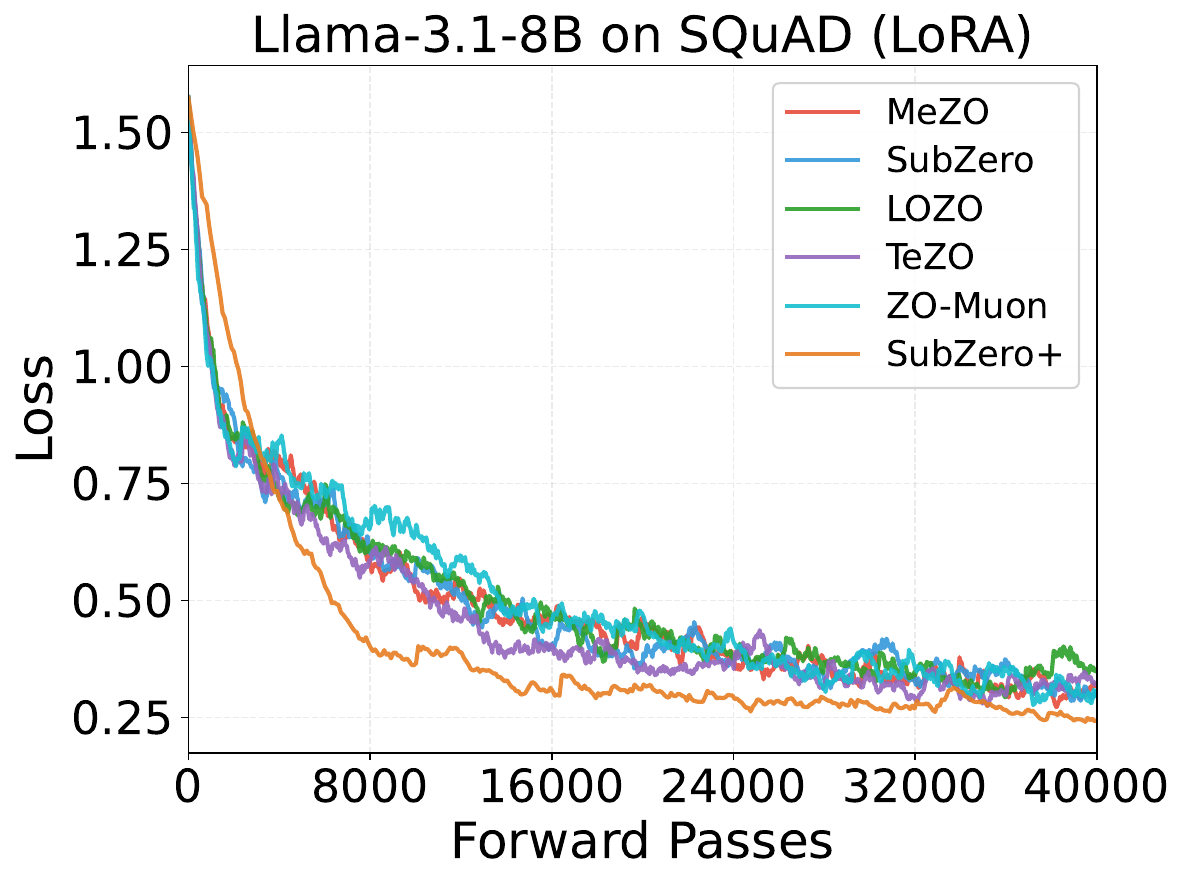}
		\par\smallskip
		{\small (a) Llama-3.1-8B on SQuAD}
	\end{minipage}
	\hfill
	\begin{minipage}[t]{0.32\textwidth}
		\centering
		\includegraphics[width=\linewidth]{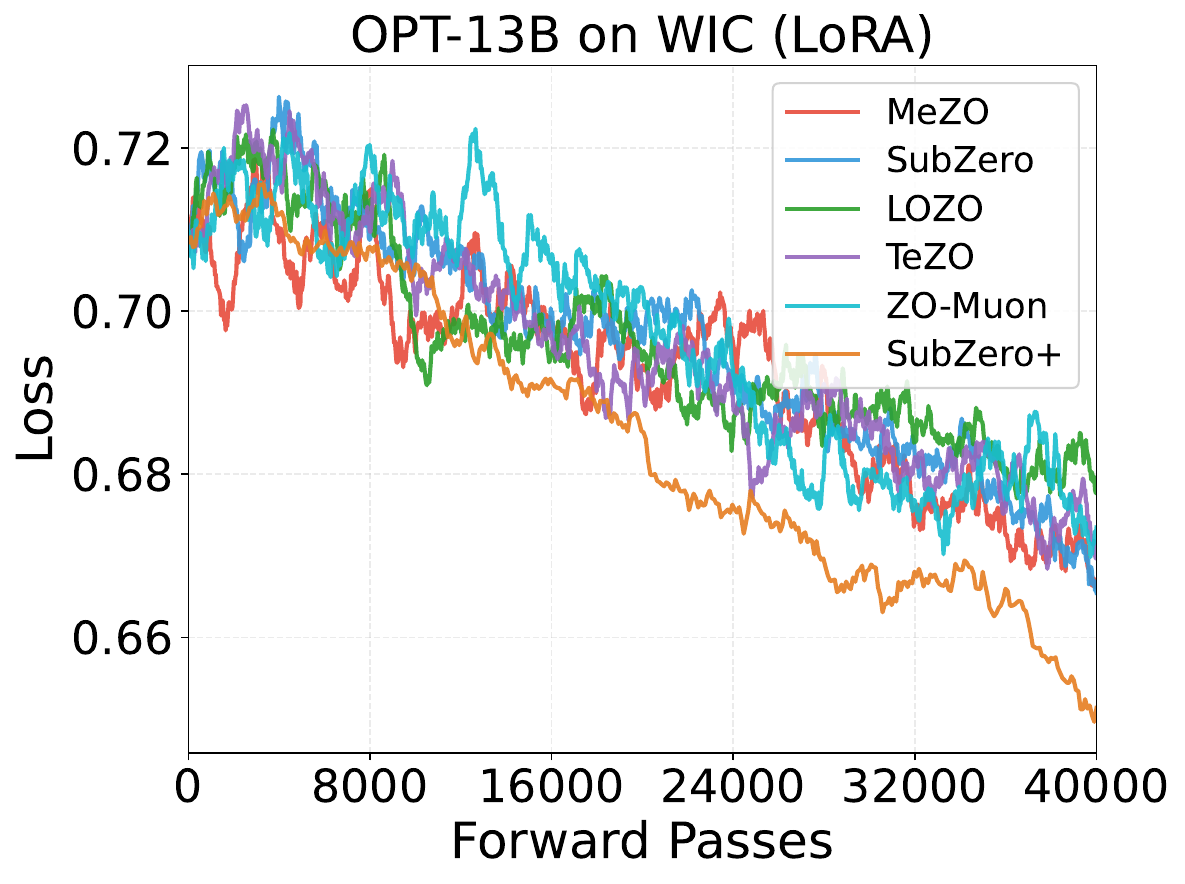}
		\par\smallskip
		{\small (b) OPT-13B on WiC}
	\end{minipage}
	\hfill
	\begin{minipage}[t]{0.32\textwidth}
		\centering
		\includegraphics[width=\linewidth]{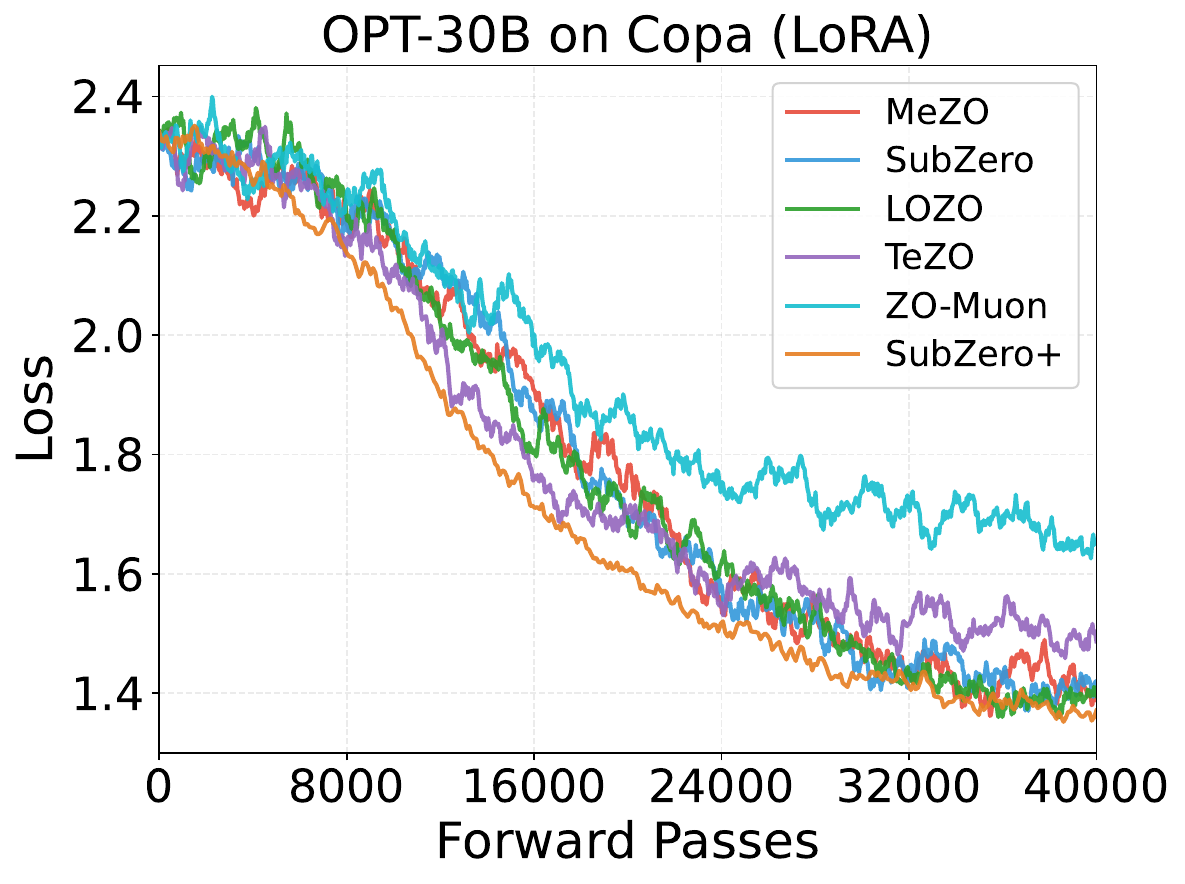}
		\par\smallskip
		{\small (c) OPT-30B on COPA}
	\end{minipage}
	\caption{Selected loss trajectories corresponding to Table~\ref{tab:lora_main}.}  
	\label{fig:lora_loss_trajectories}
	\vspace{-15pt}
\end{figure*}
\begin{table}[t]
\vspace{-15pt}
  \centering
    \caption{Performance comparison on Qwen2.5-32B. Averages are computed over five tasks.}
  \label{tab:qwen_32b}
  \renewcommand{\arraystretch}{1.0} 
  \begin{tabular}{lcccccc}
    \toprule
    Method & SST-2 & RTE & BoolQ & WIC & SQuAD& Avg. \\
    \midrule
    Zero-shot     & 61.5 & 87.0 & 84.3 & 61.8 & 85.5 & 76.0 \\
    MeZO          & 87.5 & 90.3 & 86.7 & 65.7 & 87.9 & 83.6 \\
    MeZO-LoRA     & 62.0 & 87.4 & 84.8 & 64.7 & 84.7 & 76.7 \\
    SubZero       & 89.1 & 90.3 & 87.7 & 68.8 & 89.6 & 85.1 \\
    LOZO          & 89.4 & 91.0 & 87.0 & 69.3 & \textbf{90.5} & 85.4 \\
    ZO-Muon       & 90.5 & \textbf{91.3} & \textbf{88.6} & 71.0 & \textbf{90.5} & 86.4 \\
    SubZero+ & \textbf{93.4} & \textbf{91.3} & 88.5 & \textbf{73.5} & 90.3 & \textbf{87.4} \\
    \bottomrule
  \end{tabular}
 \vspace{-6pt}  
\end{table}

\subsection{Experimental Setup}
\label{sec: experiment-setup}

\noindent\textbf{Comprehensive Validation.}
We evaluate autoregressive decoder-only LLMs, including OPT-1.3B, OPT-13B, OPT-30B~\citep{zhang2022opt}, LLaMA3.1-8B~\citep{ai2024llama3}, and Qwen2.5-32B~\citep{qwen2024},  spanning model sizes from 1.3B to 32B parameters. Tasks span the SuperGLUE benchmark~\citep{wang2019superglue}, covering classification, multiple-choice, and generation tasks. We study both the full-parameter tuning (FT) and LoRA ($r=8$ and $\alpha=16$) schemes. We compare SubZero+ against MeZO~\citep{malladi2023fine}, SubZero~\citep{yu2025zeroth}, LOZO~\citep{chen2024enhancing}, TeZO~\citep{sun2025tezo}, and ZO-Muon~\citep{lang2026powering}. For FO reference, we include Adam under FT. Zero-shot inference serves as a lower bound. MUZO~\citep{peng2025muzo}, the most directly comparable full-space multi-query baseline, is analyzed separately due to its substantially higher memory cost from maintaining full-dimensional optimizer states. Detailed memory comparisons are given in Figure~\ref{fig:fullspace_adam_memory} of Appendix~\ref{appendix:results}.

\noindent\textbf{Hyperparameters.}
SubZero+ employs default settings: perturbation queries $K=99$ and Adam parameters $\beta_1=0.9$, $\beta_2=0.95$, $\epsilon_{\mathrm{adam}}=10^{-8}$. The subspace rank $r$ and update frequency $F$ are given in Appendix~\ref{appendix:additional_results}; learning rates and method-specific structural hyperparameters are tuned by grid search per benchmark. All ZO methods use batch size 16. For every method, we select the lowest-validation-loss checkpoint within each candidate run and then choose the hyperparameter configuration with the lowest validation loss~\citep{malladi2023fine}. All methods operate under a matched 40K forward-evaluation
budget: MeZO/SubZero/LOZO/TeZO use 2 evaluations per step 
for 20K steps; ZO-Muon uses 5 evaluations per step for 8K steps; SubZero+ uses $K+1$ 
evaluations per step with 
$K=99$ and 400 steps, totaling 
40,000 forward evaluations.

\subsection{Main Results}

\subsubsection{Auto-Regressive LLMs}

We first evaluate on three autoregressive models  of varying scales: OPT-1.3B, LLaMA3.1-8B, and OPT-13B. Following prior ZO work~\citep{malladi2023fine, zhao2024second}, we use the FT scheme for all ZO methods. Results are presented in Table~\ref{tab:autoregressive_main}. SubZero+ attains the best average performance across all three models, and its advantage grows as model size increases. ZO-Muon also yields competitive results. Representative loss trajectories are shown in Figure~\ref{fig:loss_curves_table1}.

We further evaluate SubZero+ under the LoRA scheme ($r=8$, $\alpha=16$) on OPT-13B,  LLaMA3.1-8B, and OPT-30B. LoRA learns low-rank adaptation matrices for attention projections, and is widely used in practice.
Results are presented in Table~\ref{tab:lora_main}. SubZero+ achieves the best overall average on every model, with a particularly clear advantage. ZO-Muon's FT advantage fails to transfer to LoRA, likely because its matrix-sign update is designed for full-rank, near-square matrices rather than the low-rank, highly rectangular factors inherent to LoRA, whereas SubZero+ employs the reshaped subspace construction outlined in Section~\ref{sec:implementation}.
Representative convergent LoRA loss trajectories for three model--task pairs are provided in Figure~\ref{fig:lora_loss_trajectories}.

\subsubsection{Scaling to Qwen2.5-32B}

To validate SubZero+ on a substantially larger model, we additionally evaluate on Qwen2.5-32B~\citep{qwen2024} under the FT scheme. We compare against the zero-shot baseline, MeZO, MeZO with LoRA, SubZero, LOZO, and ZO-Muon. Results are presented in Table~\ref{tab:qwen_32b}. SubZero+ achieves the highest complete 5-task average.

\subsection{Ablations}
\label{sec:ablation}

\noindent\textbf{Memory Analysis.} Table~\ref{tab:memory_combined} shows that SubZero+ adds only \(\sim 1\%\) peak-memory overhead over MeZO for OPT-13B under FT, and has zero overhead under LoRA because the trainable LoRA parameters dominate.

\noindent\textbf{Learning-Rate Robustness.} Figure~\ref{fig:lr_robustness} in Appendix~\ref{appendix:results} demonstrates that SubZero+ exhibits substantially broader learning-rate tolerance than MeZO, significantly easing hyperparameter tuning in practice.

\begin{table}[t]
\vspace{-1pt}
\centering
\caption{Peak GPU memory on OPT-13B (GB, batch size 16, fp16). }
 
\label{tab:memory_combined}
\begin{tabular}{l cccc cccc}
\toprule
& \multicolumn{4}{c}{FT} & \multicolumn{4}{c}{LoRA} \\
\cmidrule(lr){2-5} \cmidrule(lr){6-9}
Task & MeZO & SubZero & SubZero+ & $\Delta$\% & MeZO & SubZero & SubZero+ & $\Delta$\% \\
\midrule
RTE   & 30.4 & 30.5 & 30.7 & +1.0 & 30.4 & 30.4 & 30.4 & +0.0 \\
BoolQ & 34.1 & 34.2 & 34.5 & +1.2 & 34.1 & 34.1 & 34.1 &  0.0 \\
SQuAD & 30.8 & 30.9 & 31.1 & +1.0 & 30.8 & 30.8 & 30.8 & +0.0 \\
DROP  & 50.4 & 50.5 & 50.8 & +0.8 & 50.4 & 50.4 & 50.4 &  0.0 \\
\bottomrule
\end{tabular}
\end{table}

\noindent\textbf{Query Count.} Under a fixed 30,000-forward budget, $K$ trades per-query variance reduction ($\propto 1/K$) against the number of steps. Table~\ref{tab:query_ablation} in Appendix~\ref{appendix:results} shows the optimal $K$ is task-dependent, so improved per-update moments do not universally favor larger $K$.

\section{Conclusion}
\label{sec: conclusion}
\vspace*{-0.5em} 

SubZero+ enables adaptive ZO optimization through dual low-dimensionality: multi-query estimation in random subspaces and Adam updates in low-dimensional spaces. In the exact-directional limit, we show that the $K$-query coefficient estimator is conditionally unbiased and that its covariance, mean-squared error, and query-induced second-moment inflation decrease exactly as $1/K$. Experimentally, SubZero+ achieves consistent gains over ZO baselines across model scales from 1.3B to 32B parameters under full-parameter and LoRA schemes, with minimal memory cost.

\noindent\textbf{Limitations and Future Work.} Key limitations are data-independent subspace construction and momentum resets that discard temporal context. Future work should explore data-driven basis selection and moment transport across basis refreshes.

\section*{Acknowledgments}
This work is supported by the NSF of China (grant no. 62437001), Xiaomi 2025 Open Research Project, the Ministry of Education, Singapore, under its AcRF Tier 2 Funding (Proposal ID: T2EP20224-0048). Any opinions, findings and conclusions or recommendations expressed in this material are those of the author(s) and do not reflect the views of the Ministry of Education, Singapore.
 
\bibliography{main}
\bibliographystyle{arxiv_public}

\newpage
\appendix
\section{Appendix}

\subsection{Additional Experimental Results}
\label{appendix:results}

\begin{figure}[h]
	\centering
	\subfigure[Loss Landscape.]{
		\includegraphics[width=0.330\linewidth]{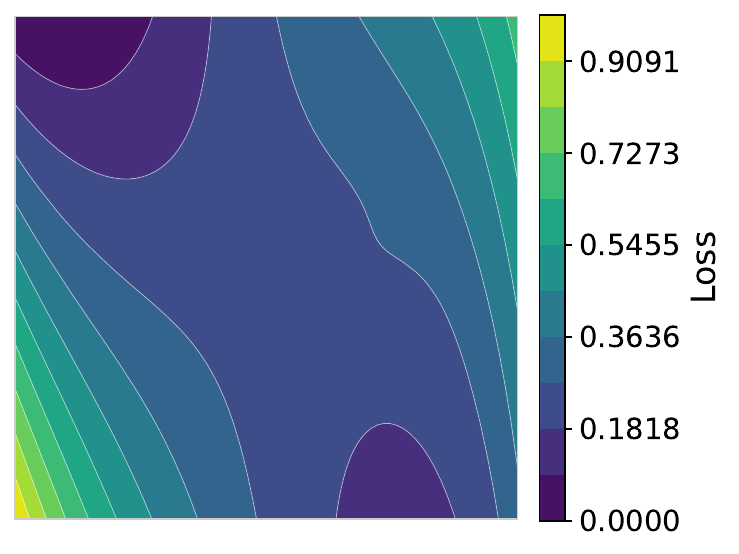}
		\label{fig:gl_a}
	}
	\hfill
	\subfigure[SubZero ($\|\overline{\mZ}\|$, $K=1$).]{
		\includegraphics[width=0.305\linewidth]{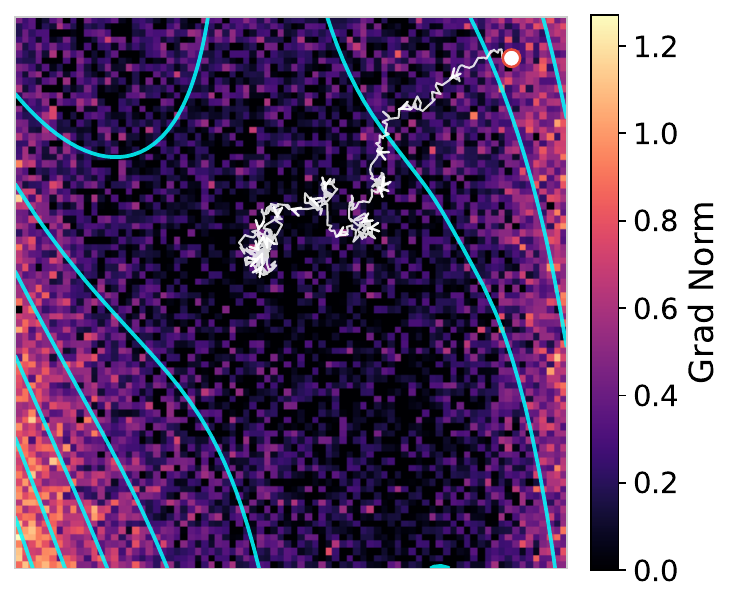}
		\label{fig:gl_b}
	}
	\hfill
	\subfigure[SubZero+ ($\|\overline{\mZ}\|$, $K=100$).]{
		\includegraphics[width=0.305\linewidth]{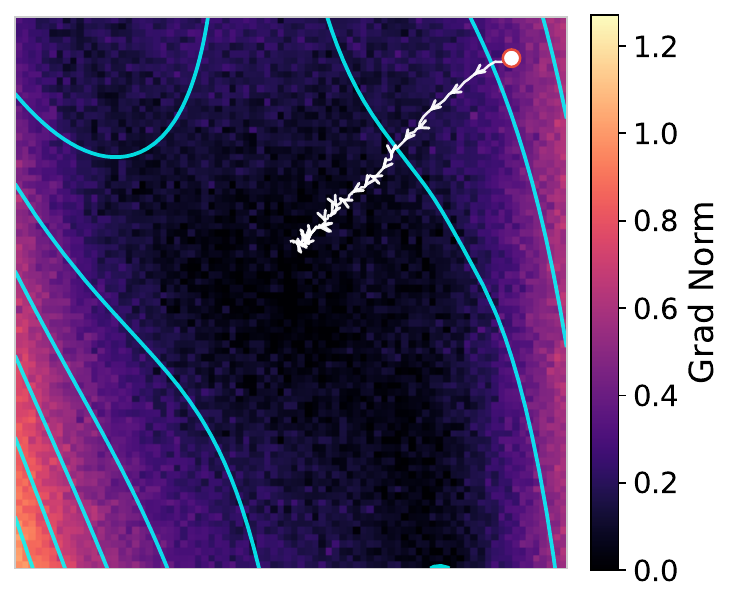}
		\label{fig:gl_c}
	}
		\caption{Multi-query estimation reduces gradient noise. (a) A 2D loss surface with steep walls and a flat valley. (b) SubZero with $K=1$. The estimated gradient-norm field is noisy. (c) SubZero+ with $K=100$. Averaging produces a visibly smoother gradient-norm field. All panels share the same domain and starting point. 
		}
		\label{fig:grad_norm_landscape}
\end{figure}

\noindent\textbf{Motivation of Multi-Query.} We employ a simple 2D loss surface to demonstrate the motivation for SubZero+'s use of multi-query. The results are presented in Figure~\ref{fig:grad_norm_landscape}.

\begin{figure}[t]
	\centering
	\includegraphics[width=0.72\linewidth]{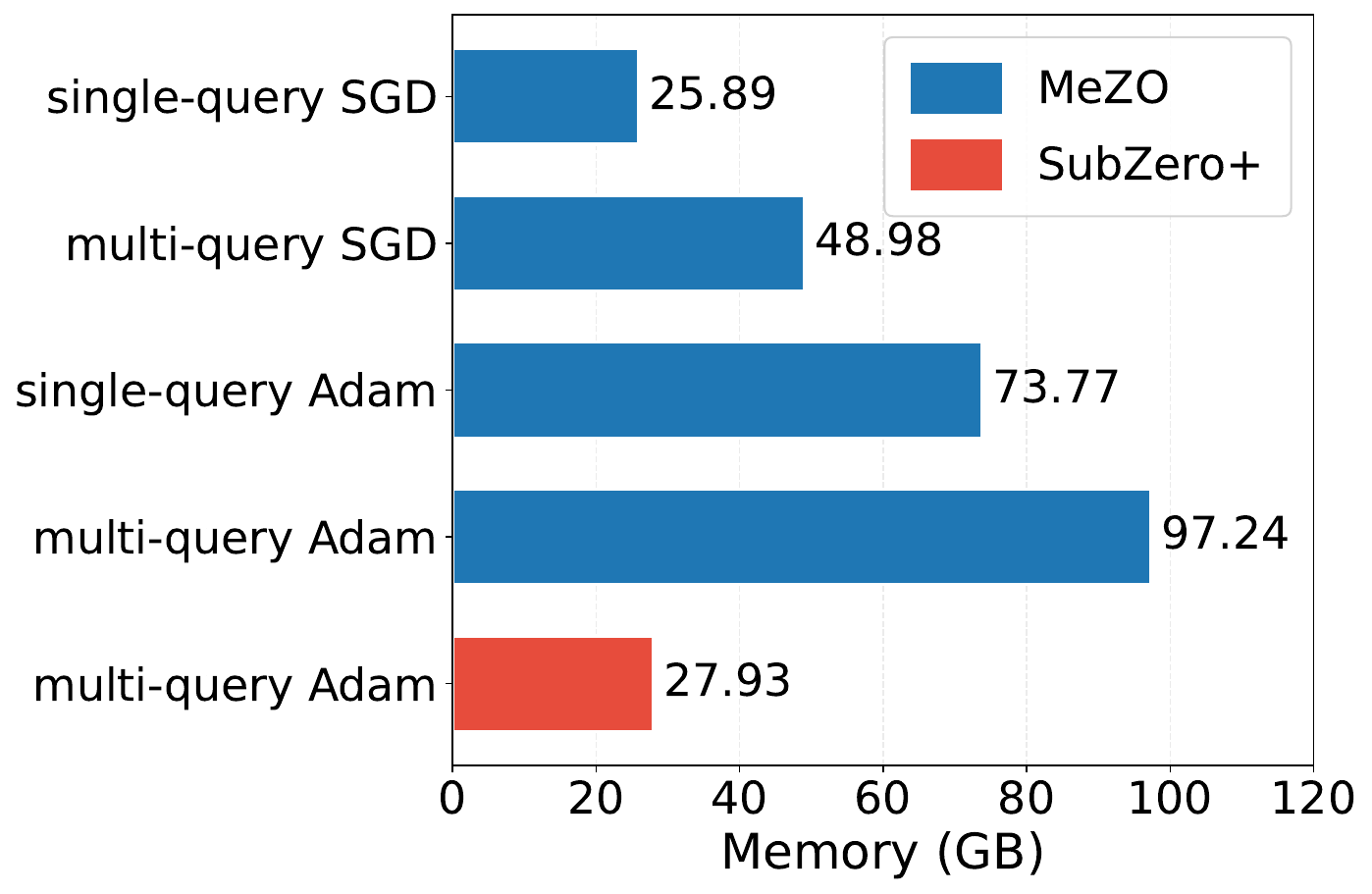}
	\caption{Peak GPU memory for full-parameter tuning of OPT-13B on SST-2 (batch size 16, fp16). The blue bars apply single- or multi-query SGD/Adam in full parameter space on top of MeZO; the red bar is SubZero+ with multi-query coefficient-space Adam. The 97.24~GB bar represents straightforward full-space query accumulation rather than a direct reproduction of MUZO's layer-wise implementation.}
	\label{fig:fullspace_adam_memory}
\end{figure}

\noindent\textbf{Memory Cost of Full-Space Multi-Query Adam.}
MUZO combines multi-query averaging with Adam in the full parameter space~\citep{peng2025muzo}. Its layer-wise seed schedule reconstructs perturbations on demand and accumulates the multi-query estimate one layer at a time. Thus, MUZO retains a current-layer gradient buffer but avoids storing all query-wise full gradients simultaneously; its persistent first- and second-moment buffers nevertheless remain $O(d)$. Accordingly, MUZO reports 35.2~GB for full-parameter MUZO-Adam versus 13.6~GB for MeZO-SGD on OPT-2.7B/SST-2, and introduces QAdam to reduce this optimizer-state cost. Our OPT-13B measurement in Figure~\ref{fig:fullspace_adam_memory} separates the same bottleneck: MeZO-style full-space Adam uses 73.77~GB even with a single query, while straightforward multi-query accumulation raises peak memory to 97.24~GB, exceeding a single 80~GB accelerator. SubZero+ instead stores the accumulated estimate and both moments in coefficient space and uses 27.93~GB.

\noindent\textbf{Ablations on Query Count.} 
Ablation study on perturbation-query count $K$ across tasks under FT scheme. Results on OPT-1.3B with SubZero+ (sign-corrected QR subspaces, $r=24$) across SST-2, RTE, and MultiRC. The perturbation-query budget is fixed at 30,000 for all $K$; larger $K$ uses proportionally fewer steps, and each step additionally evaluates one baseline loss. All experiments use full-parameter tuning. The best accuracy for each task is bolded, and the best overall $K$ for the average is bolded in the Avg.\ column.

\begin{table*}[h]
\vspace{-5pt}
\caption{Ablation study on perturbation-query count $K$ across tasks under FT scheme. Results on OPT-1.3B with SubZero+ (sign-corrected QR subspaces, $r=24$) across SST-2, RTE, and MultiRC.}
\vspace{2pt}
\centering
\resizebox{\textwidth}{!}{
\begin{tabular}{lcccccccccccc}
	\toprule
	Dataset & 1 & 5 & 10 & 20 & 30 & 50 & 70 & 80 & 100 & 150 & 200 & $\Delta$ \\
	\midrule
	SST-2   & 50.9 & 52.6 & 56.7 & 69.8 & 88.3 & 90.9 & 91.5 & 91.6 & \textbf{92.2} & 89.9 & 89.7 & +41.3 \\
	RTE     & 47.7 & 52.7 & 49.8 & 57.0 & 62.8 & \textbf{63.9} & 58.1 & 60.3 & 57.0 & 61.4 & 61.0  & +9.3\\
	MultiRC & 55.5 & 55.5 & 52.3 & 59.8 & 57.6 & 60.0 & 57.6 & 59.3 & 59.6 & 58.7 & \textbf{61.1} & +4.1\\
	\midrule
	Avg.     & 51.4 & 53.6 & 52.9 & 62.2 & 69.6 & \textbf{71.6} & 69.3 & 70.4 & 69.6 & 70.0 & 70.6  & +18.2 \\
	\bottomrule
\end{tabular}
}
\vspace{-5pt}
\label{tab:query_ablation}
\end{table*}

\begin{figure}[tbp]
	\centering
	\begin{minipage}[t]{0.48\textwidth}
		\centering
		\includegraphics[width=\linewidth]{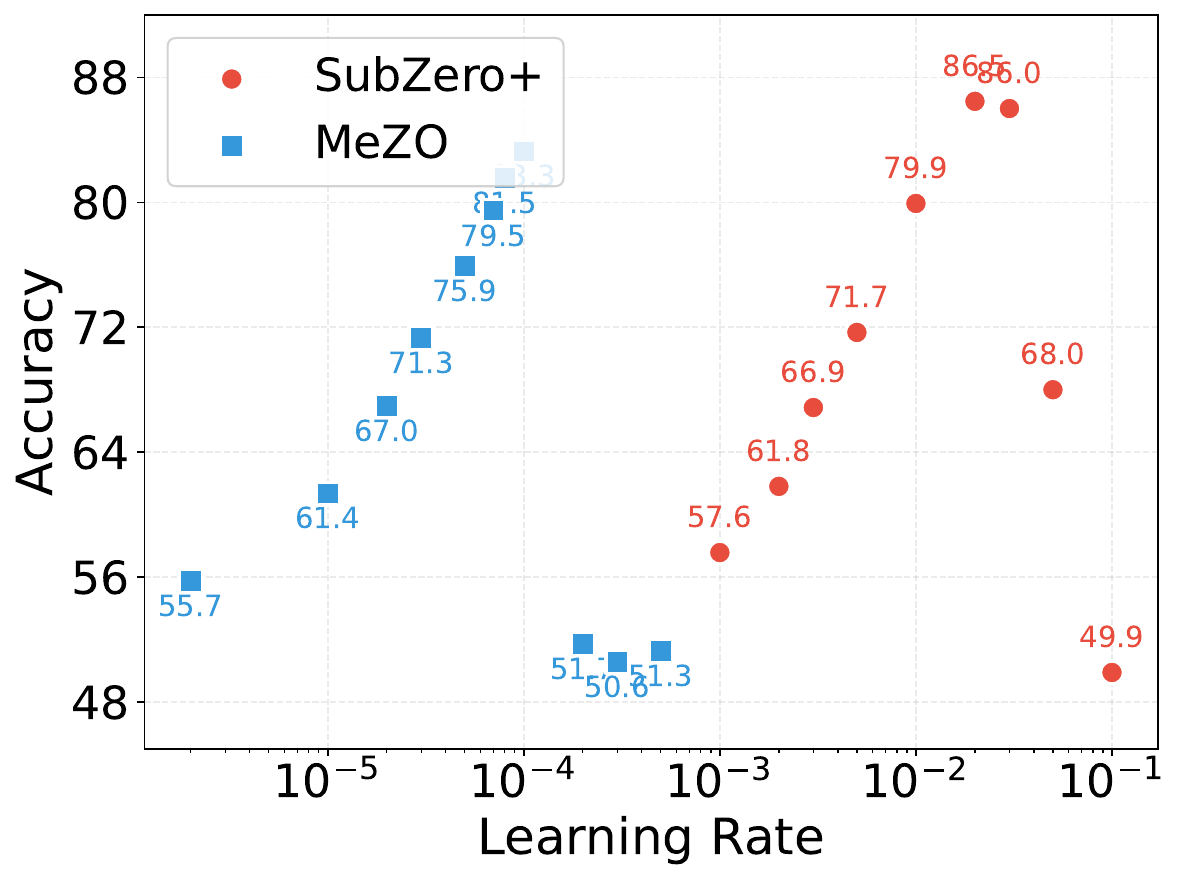}
		\footnotesize{(a) Varying learning rates}
	\end{minipage}
	\hfill
	\begin{minipage}[t]{0.48\textwidth}
		\centering
		\includegraphics[width=\linewidth]{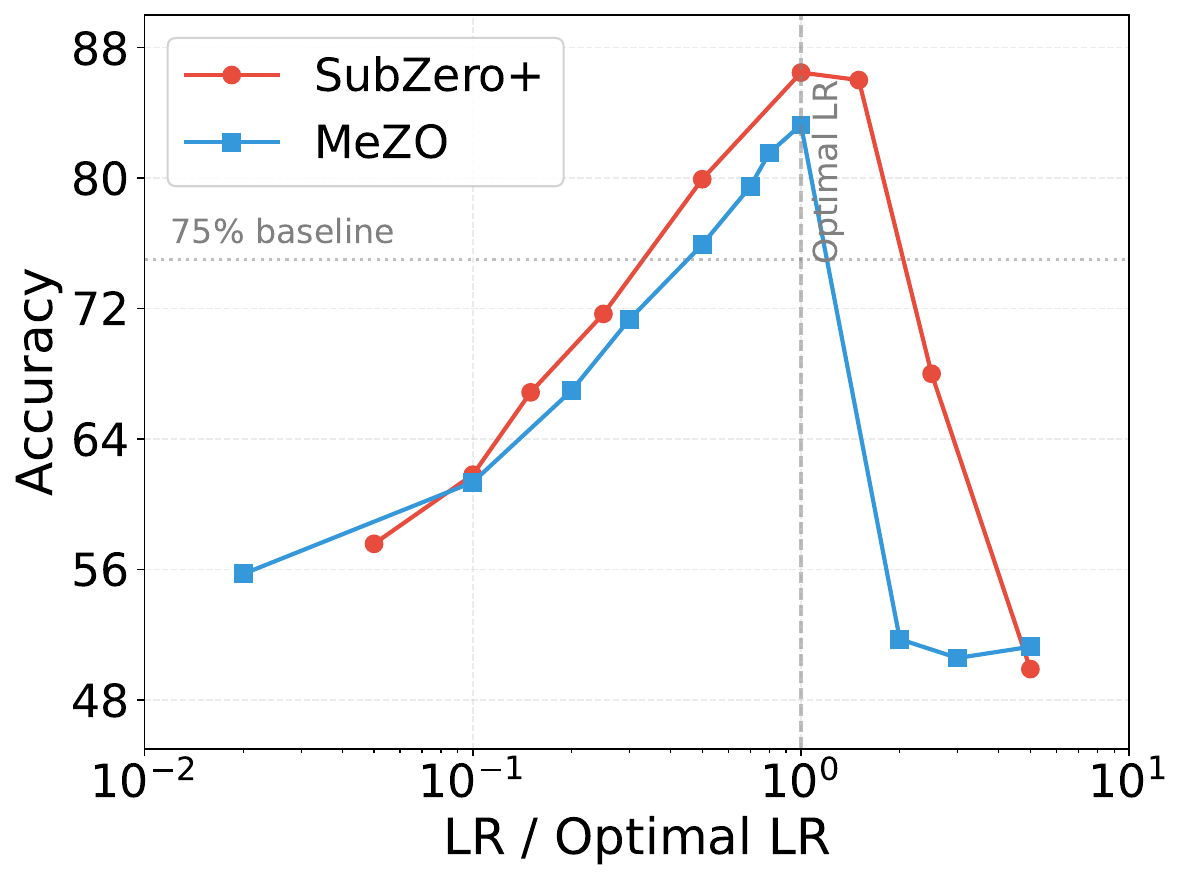}
		\footnotesize{(b) Normalized performance}
	\end{minipage}
	\caption{Learning-rate robustness of SubZero+ and MeZO for LoRA of Qwen3-0.6B~\citep{yang2025qwen3technicalreport} on SST-2. (a) Training accuracy versus learning rate (LR). SubZero+ maintains high accuracy over a wide LR range, whereas MeZO degrades quickly outside its optimal value. (b) Normalized results, with each LR scaled by its optimum. SubZero+ shows a symmetric plateau, while MeZO collapses when LR is larger than the optimum. }  
	\label{fig:lr_robustness}
    \vspace{-1em}
\end{figure}

\noindent\textbf{Learning-Rate Robustness.}
We empirically compare the learning-rate (LR) robustness of the complete SubZero+ and MeZO methods by fine-tuning Qwen3-0.6B~\citep{yang2025qwen3technicalreport} with LoRA on SST-2 and sweeping learning rates across four orders of magnitude. As shown in Figure~\ref{fig:lr_robustness}(a), SubZero+ achieves the highest training accuracy of $86.5\%$ at LR $=2\times 10^{-2}$, and retains above $79.0\%$ accuracy across a $3\times$ LR range ($10^{-2}\sim 3\times 10^{-2}$). In comparison, MeZO reaches its best accuracy of $83.3\%$ at LR $=1\times 10^{-4}$, but its performance window is considerably narrower. When LR is doubled to $2\times 10^{-4}$, MeZO's accuracy drops by over $31$ percentage points to $51.7\%$, whereas SubZero+ drops by only $0.5$ percentage points when LR is increased by $1.5\times$ to $3\times 10^{-2}$. This comparison does not isolate the effect of multi-query averaging from the other differences between the methods.

To eliminate the scale difference in optimal LRs, we normalize each method's LR by its own optimum in Figure~\ref{fig:lr_robustness}(b). The normalized view reveals that SubZero+ exhibits a broad, symmetric performance plateau around the optimal LR, with a standard deviation of merely $2.98\%$ within the $[\text{opt}/2, 2\times\text{opt}]$ interval. MeZO, on the other hand, shows highly asymmetric behavior: its accuracy degrades gracefully when LR is below the optimum but collapses abruptly when LR is even slightly above it. This suggests that SubZero+ is substantially more robust to learning rate selection in practical LoRA fine-tuning scenarios, where the exact optimal LR is often unknown a priori.

\subsection{Experimental Details}
\label{appendix:additional_results}

\noindent\textbf{Implementation Details.} For one-dimensional parameters, SubZero uses MeZO with single-query SGD, whereas SubZero+ uses multi-query Adam. These scalar parameters are outside the Haar--Kronecker theory in Section~\ref{sec:theory}. SubZero+ inherits the random-seed trick and per-layer update strategy from SubZero~\citep{yu2025zeroth} for memory-efficient perturbation regeneration. For LoRA, also following SubZero, SubZero+ reshapes each two-dimensional trainable factor into an approximately square matrix with the same number of elements before constructing the projection bases; each perturbation and update is reshaped back to the original parameter shape before being applied. The projection matrices $\mU_i, \mV_i$ are regenerated every $F$ steps and cached in between. Upon each regeneration, the moment buffers $\mM_i, \mS_i$ are reset to zero and their local bias-correction counter restarts. Each matrix-valued layer contributes $2r^2$ moment floats; first- and second-moment states for separately handled one-dimensional parameters are counted in addition. The cached bases require separate storage of $\sum_i (m_i+n_i)r$ floats. Multi-query perturbations are generated sequentially; each is regenerated from its seed before the forward pass and immediately subtracted afterward, keeping peak memory at inference levels.

\noindent\textbf{Forward-Pass and Perturbation-Query Accounting.}
For the main comparison, MeZO and SubZero use two forward passes per step for 20K steps, for a total of 40K forward passes. SubZero+ uses $K$ perturbation queries and one unperturbed baseline forward pass per step. With the default $K=99$ and 400 steps, it uses $99\times400=39{,}600$ perturbation-query forward passes and 400 baseline forward passes, again totaling 40K forward passes. Thus, the main comparison matches total forward passes, while $K$ always denotes the number of perturbation queries rather than the total number of forward passes per update. The query-count ablation likewise fixes a total forward-pass budget and counts the additional baseline forward pass in every update.

\noindent\textbf{Hyperparameter Configurations.}
All experiments are conducted on a single NVIDIA H200 (141GB) GPU, except for OPT-1.3B which is run on a single NVIDIA RTX 4090 (24GB) GPU. We use PyTorch 2.1.0 with CUDA 12.1.

The method-specific hyperparameter ranges are initialized from the corresponding original papers and adapted to our model suite and common 40K-forward-pass budget. Tables~\ref{tab:hyper_ft} and~\ref{tab:hyper_lora} report the exact search grids and fixed configurations used in our experiments. Unless otherwise specified, all ZO methods use batch size 16, perturbation scale $10^{-3}$, a constant learning-rate schedule, and zero weight decay. MeZO, SubZero, LOZO, and TeZO use central differences with two forward passes per optimization step and are trained for 20,000 steps. ZO-Muon uses four forward-difference perturbation queries and one shared baseline forward pass per step for 8,000 steps. SubZero+ uses 99 forward-difference perturbation queries and one shared baseline forward pass per step for 400 steps. Thus, every method uses 40,000 training-time forward passes. For every method, we select the lowest-validation-loss checkpoint within each candidate run and then choose the learning rate and method-specific structural hyperparameters by validation loss.

\begin{table}[ht]
		\caption{Hyperparameter configurations and search grids for the FT scheme. Common settings are batch size 16, perturbation scale $10^{-3}$, constant learning rate, and zero weight decay.}
	\label{tab:hyper_ft}
	\centering
		\scriptsize
		\renewcommand{\arraystretch}{1.08}
		\begin{tabular}{@{}p{0.20\columnwidth}p{0.76\columnwidth}@{}}
			\toprule
			Method & Configuration \\
			\midrule
			MeZO
			& learning rate $\{10^{-7},5\times10^{-7},10^{-6}\}$; 20,000 steps (2 forward passes/step) \\
			\midrule
			SubZero
			& learning rate $\{10^{-7},5\times10^{-7},10^{-6}\}$; rank $r=16$ (OPT-1.3B) or $128$ (others); update frequency $F\in\{500,1000,2000\}$; 20,000 steps (2 forward passes/step) \\
			\midrule
			LOZO
			& learning rate $\{10^{-7},5\times10^{-7},10^{-6}\}$; rank $r=8$; reuse interval $\nu=100$; 20,000 steps (2 forward passes/step) \\
			\midrule
			TeZO
			& learning rate $\{10^{-7},5\times10^{-7},10^{-6}\}$; fixed rank $r=64$; 20,000 steps (2 forward passes/step) \\
			\midrule
			ZO-Muon
			& learning rate $\{3\times10^{-3},10^{-2},3\times10^{-2}\}$; projection rank $r\in\{64,128\}$; $N_q=4$; projection interval 100; 5 Newton--Schulz iterations; 8,000 steps (4 perturbation queries + 1 baseline forward pass/step) \\
			\midrule
			SubZero+
			& learning rate $\{10^{-3},5\times10^{-3},10^{-2}\}$; rank $r=16$ (OPT-1.3B) or $128$ (others); update frequency $F\in\{50,100\}$; $K=99$; 400 steps (99 perturbation queries + 1 baseline forward pass/step); Adam $(\beta_1,\beta_2)=(0.9,0.95)$ and $\epsilon_{\mathrm{adam}}=10^{-8}$ \\
			\bottomrule
		\end{tabular}
	\vspace{-5pt}
\end{table}

\begin{table}[ht]
		\caption{Hyperparameter configurations and search grids for the LoRA scheme. LoRA rank is 8 and $\alpha=16$ for all methods; common ZO settings are as in Table~\ref{tab:hyper_ft}.}
	\label{tab:hyper_lora}
	\centering
		\scriptsize
		\renewcommand{\arraystretch}{1.08}
		\begin{tabular}{@{}p{0.20\columnwidth}p{0.76\columnwidth}@{}}
			\toprule
			Method & Configuration \\
			\midrule
			MeZO
			& learning rate $\{1.5\times10^{-5},3\times10^{-5},5\times10^{-5}\}$; 20,000 steps (2 forward passes/step) \\
			\midrule
			SubZero
			& learning rate $\{1.5\times10^{-5},3\times10^{-5},5\times10^{-5}\}$; rank $r\in\{4,8,16\}$; update frequency $F\in\{500,1000,2000\}$; 20,000 steps (2 forward passes/step) \\
			\midrule
			LOZO
			& learning rate $\{1.5\times10^{-5},3\times10^{-5},5\times10^{-5}\}$; rank $r=8$; reuse interval $\nu=100$; 20,000 steps (2 forward passes/step) \\
			\midrule
			TeZO
			& learning rate $\{1.5\times10^{-5},3\times10^{-5},5\times10^{-5}\}$; fixed rank $r=64$; 20,000 steps (2 forward passes/step) \\
			\midrule
			ZO-Muon
			& learning rate $\{10^{-3},3\times10^{-3},10^{-2}\}$; projection rank $r\in\{8,16\}$ (selected: 16); $N_q=4$; projection interval 100; 5 Newton--Schulz iterations; 8,000 steps (4 perturbation queries + 1 baseline forward pass/step) \\
			\midrule
			SubZero+
			& learning rate $\{5\times10^{-3},10^{-2},5\times10^{-2}\}$; rank $r\in\{4,8,16\}$; update frequency $F\in\{50,100\}$; $K=99$; 400 steps (99 perturbation queries + 1 baseline forward pass/step); Adam $(\beta_1,\beta_2)=(0.9,0.95)$ and $\epsilon_{\mathrm{adam}}=10^{-8}$ \\
			\bottomrule
		\end{tabular}
		\vspace{-5pt}
\end{table}

\noindent\textbf{Baseline Implementation Details.}
For LOZO~\citep{chen2024enhancing}, we use its low-rank central-difference estimator with rank $r=8$: one Gaussian factor is reused for $\nu=100$ consecutive steps, while the complementary factor is resampled at every step. Our implementation normalizes each perturbation direction by $\sqrt{r}$; therefore, the learning rates in Tables~\ref{tab:hyper_ft} and~\ref{tab:hyper_lora} are the actual step sizes passed to our implementation rather than the $\alpha/r$ convention used in the original LOZO tables.

For TeZO~\citep{sun2025tezo}, no publicly available official implementation could be located at the time of our experiments, so we implemented Algorithm~1 of the original paper. For every two-dimensional trainable parameter, the spatial factors $U$ and $V$ are sampled once at initialization and shared throughout training, while a temporal coefficient vector is sampled independently at every optimization step. We use the fixed-rank configuration $r=64$; one-dimensional parameters use full-space Gaussian perturbations. Because the spatial factors remain fixed throughout training, TeZO has no periodic subspace-refresh hyperparameter in our implementation.

For ZO-Muon~\citep{lang2026powering}, the FT experiments follow the original choices of four forward-difference queries, projection ranks $\{64,128\}$, projection interval 100, and Newton--Schulz matrix-sign approximation, while broadening the reported learning rate $10^{-2}$ to a three-point grid. The original LOZO, TeZO, and ZO-Muon papers do not report the same ZO-on-LoRA setting used in our LoRA experiments. We therefore apply their estimators to the trainable LoRA matrices while freezing the backbone. In particular, the ZO-Muon projection rank is distinct from the LoRA adapter rank; we search $\{8,16\}$ and select 16. These LoRA results are extensions implemented in our common framework rather than direct reproductions of source-paper LoRA results.

\subsection{Proofs for Section~\ref{sec:theory}}
\label{appendix:theory}

All expectations below are conditional on $\mathcal G$. Thus $\va$, $\vw$, $\mP$, and $f_{\mathcal B}$ are fixed, while $\vz_1,\ldots,\vz_K$ are independent standard Gaussian vectors.

\subsubsection{Proof of Theorem~\ref{thm:first_moment}}

\begin{proof}
For one exact-directional query, define
\begin{align}
\vx&=\vz\vz^\top\va.
\end{align}
Isserlis' identity yields
\begin{align}
\mathbb E\!\left[\vx\mid\mathcal G\right]&=\va,
&
\mathbb E\!\left[\vx\vx^\top\mid\mathcal G\right]
&=\|\va\|^2\mI_q+2\va\va^\top.
\label{eq:single_query_moments}
\end{align}
Therefore,
\begin{align}
\operatorname{Cov}(\vx\mid\mathcal G)
=\va\va^\top+\|\va\|^2\mI_q.
\end{align}
Averaging $K$ conditionally independent copies preserves the mean and divides
the covariance by $K$. Taking the trace of the resulting covariance proves the
mean-squared-error identity in Eq.~(\ref{eq:exact_directional_error}).
\end{proof}

\subsubsection{Proof of Theorem~\ref{thm:second_moment}}

\begin{proof}
The $p$th diagonal entry of Eq.~(\ref{eq:exact_directional_error}) gives
\begin{align}
\operatorname{Var}[\widehat a_p\mid\mathcal G]
=\frac1K\left(a_p^2+\|\va\|^2\right).
\end{align}
Combining this with
$\mathbb E[\widehat a_p\mid\mathcal G]=a_p$ yields
\begin{align}
\mathbb E\!\left[\widehat a_p^2\mid\mathcal G\right]
=a_p^2+\operatorname{Var}[\widehat a_p\mid\mathcal G]
=\left(1+\frac1K\right)a_p^2+\frac{\|\va\|^2}{K}.
\end{align}
Stacking the coordinate identities proves Eq.~(\ref{eq:second_moment_identity}).
\end{proof}

\end{document}